\documentclass[pdflatex,referee,sn-nature]{sn-jnl} 

\usepackage{hyperref}
\usepackage{xr-hyper}
\usepackage{docmute}  

\usepackage[version=4]{mhchem}
\usepackage{graphicx}%
\usepackage{multirow}%
\usepackage{amsmath,amssymb,amsfonts}%
\usepackage{amsthm}%
\usepackage{mathrsfs}%
\usepackage[title]{appendix}%
\usepackage{xcolor}%
\usepackage{textcomp}%
\usepackage{manyfoot}%
\usepackage{booktabs}%
\usepackage{algorithm}%
\usepackage{algorithmicx}%
\usepackage{algpseudocode}%
\usepackage{listings}%
\usepackage{tabularx,booktabs}
\newcolumntype{Y}{>{\centering\arraybackslash}X}

\usepackage{xr-hyper}
\usepackage{hyperref}

\usepackage{color}
\definecolor{teal}{rgb}{0.0, 0.5, 0.5}

\usepackage[acronym]{glossaries}

\newcommand{\figtoc}[1]{\addcontentsline{toc}{subsection}{Fig.~\thefigure.\ #1}}
\newcommand{\tabtoc}[1]{\addcontentsline{toc}{subsection}{Table~\thetable.\ #1}}

\theoremstyle{thmstyleone}%
\theoremstyle{thmstyletwo}%

\theoremstyle{thmstylethree}%

\newacronym{n2v}{N2V}{Noise2Void}
\newacronym{umvd}{UMVD}{Unsupervised Microscopy Video Denoising}
\newacronym{pde}{PDE}{partial differential equation}
\newacronym{snr}{SNR}{signal-to-noise ratio}
\newacronym{psnr}{PSNR}{peak signal-to-noise ratio}
\newacronym{ssim}{SSIM}{structural similarity index metric}

\newcommand{\beginsupplement}{%
  \setcounter{section}{0}%
  \renewcommand{\thesection}{S\arabic{section}}%
  \renewcommand{\theHsection}{supp.\arabic{section}}%

  \setcounter{figure}{0}%
  \renewcommand{\thefigure}{S\arabic{figure}}%
  \renewcommand{\theHfigure}{supp.\arabic{figure}}%

  \setcounter{table}{0}%
  \renewcommand{\thetable}{S\arabic{table}}%
  \renewcommand{\theHtable}{supp.\arabic{table}}%

  \setcounter{equation}{0}%
  \renewcommand{\theequation}{S\arabic{equation}}%
  \renewcommand{\theHequation}{supp.\arabic{equation}}%
}
\usepackage{bibunits}
\defaultbibliography{zotero_bib_final}

\begin{document}

\begin{bibunit}[sn-nature]                 

\title[Article Title]{Deep learning denoising unlocks quantitative insights in \textit{operando} materials microscopy}


\author[1]{\fnm{Samuel} \sur{Degnan-Morgenstern}}\email{stm16109@mit.edu}

\author[1]{\fnm{Alexander E.} \sur{Cohen}}

\author[2]{\fnm{Rajeev} \sur{Gopal}}
\author[3]{\fnm{Megan} \sur{Gober}}


\author[4]{\fnm{George J.} \sur{Nelson}}
\author[2,5]{\fnm{Peng} \sur{Bai}}
\author*[1,6]{\fnm{Martin Z.} \sur{Bazant}}\email{bazant@mit.edu}



\affil*[1]{\orgdiv{Department of Chemical Engineering}, \orgname{Massachusetts Institute of Technology}, \orgaddress{\street{25 Ames St.}, \city{Cambridge}, \postcode{02139}, \state{MA}, \country{USA}}}

\affil[2]{\orgdiv{Department of Energy, Environmental and Chemical Engineering}, \orgname{Washington University in St. Louis}, \orgaddress{\street{1 Brookings Dr}, \city{St. Louis}, \state{MO}, \postcode{63130}, \country{USA}}}

\affil[3]{\orgdiv{Research Institute}, \orgname{University of Alabama in Huntsville}, \orgaddress{\city{Huntsville}, \state{AL}, \country{USA}}}

\affil[4]{\orgdiv{Department of Mechanical and Aerospace Engineering}, \orgname{University of Alabama in Huntsville}, \orgaddress{\city{Huntsville}, \state{AL}, \country{USA}}}

\affil[5]{\orgdiv{Institute of Materials Science and Engineering}, \orgname{Washington University in St. Louis}, \orgaddress{\street{1 Brookings Dr}, \city{St. Louis}, \state{MO}, \postcode{63130}, \country{USA}}}

\affil[6]{\orgdiv{Department of Mathematics}, \orgname{Massachusetts Institute of Technology}, \orgaddress{\street{182 Memorial Dr.}, \city{Cambridge}, \postcode{02139}, \state{MA}, \country{USA}}}








\abstract{
\textit{Operando} microscopy provides direct insight into the dynamic chemical and physical processes that govern functional materials, yet measurement noise limits the effective resolution and undermines quantitative analysis.
Here, we present a general framework for integrating unsupervised deep learning-based denoising into quantitative microscopy workflows across modalities and length scales.
Using simulated data, we demonstrate that deep denoising preserves physical fidelity, introduces minimal bias, and reduces uncertainty in model learning with partial differential equation (PDE)-constrained optimization.
Applied to experiments, denoising reveals nanoscale chemical and structural heterogeneity in scanning transmission X-ray microscopy (STXM) of lithium iron phosphate (LFP), enables automated particle segmentation and phase classification in optical microscopy of graphite electrodes, and reduces noise-induced variability by nearly 80\% in neutron radiography to resolve heterogeneous lithium transport.
Collectively, these results establish deep denoising as a powerful, modality-agnostic enhancement that advances quantitative \textit{operando} imaging and extends the reach of previously noise-limited techniques.
}




\maketitle

\section*{Main}\label{sec:main}

\textit{Operando} microscopy is fundamental for characterizing materials dynamics, providing direct visualization of heterogeneous processes across multiple length and time scales.
Such techniques are especially critical for energy storage materials, where physical properties depending on chemical composition evolve continuously during operation~\cite{liu_review_2019,wu_visualizing_2018,yang_operando_2021}.
Electron and synchrotron-based X-ray microscopy provide high-fidelity chemical and structural mapping with atomistic ($<$ 0.1 nm) to nanoscale resolution (10–500 nm) ~\cite{chang_quantifying_2023,he_operando_2018,holtz_nanoscale_2014,lin_synchrotron_2017,lim_origin_2016,deng_correlative_2021}, while optical methods capture dynamic heterogeneity from the single-particle (0.1-10 $\mu$m) to the electrode scale ($\sim$1mm) ~\cite{merryweather_operando_2021,harris_direct_2010,agrawal_operando_2021,agrawal_dynamic_2022,lu_multiscale_2023}.
Neutron imaging extends this reach to the device scale, probing macroscopic transport and degradation with exceptional sensitivity to light elements such as lithium \cite{ziesche_neutron_2022,gober_high_2025}.
Across all these modalities, measurement noise is a significant challenge, as it obscures subtle spatiotemporal variations and complicates both qualitative interpretation and quantitative analysis \cite{merryweather_operando_2021,bak_situoperando_2018,perez_neutron_2023,ziesche_4d_2020}.
In practice, researchers must balance signal quality, spatiotemporal resolution, and noise: driving costly trade-offs in beamline experiments and constraining high-throughput, pixel-resolved analysis in benchtop microscopy~\cite{zhang_visualizing_2025}.

Deep learning–based denoising has emerged as a state-of-the-art solution for addressing noise in microscopy, achieving major advances in biological imaging by surpassing classical approaches and improving downstream analyses such as segmentation and classification \cite{laine_imaging_2021,belthangady_applications_2019,weigert_content-aware_2018,krull_noise2void_2019,lecoq_removing_2021,sheth_unsupervised_2021,aiyetigbo_unsupervised_2024,li_reinforcing_2021}.
Self-supervised strategies, which learn directly from raw data without curated labels, are particularly well-suited for enhancing image and video quality in real experimental workflows \cite{morales_evaluating_2023}.
In materials science, these methods have so far been applied mainly to electron microscopy, where they revealed hidden atomic-scale dynamics and improved segmentation in catalytic nanoparticles \cite{crozier_visualizing_2025,kim_self-supervised_2025}.
However, their application to other microscopy modalities and to larger, device-relevant length scales remains largely unexplored.
Extending these approaches to energy materials presents unique challenges, since rigorous validation is required to ensure denoising preserves quantitative fidelity for analysis and mechanistic interpretation.

\begin{figure}[H]
\centering
\includegraphics[width=\textwidth]{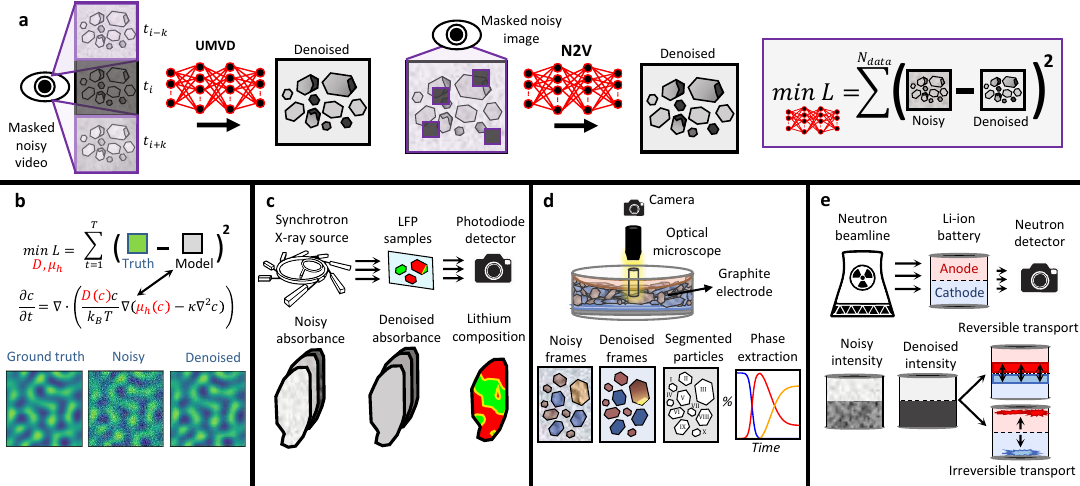}
\caption{ \textbf{Deep learning–based denoising integrated into microscopy workflows enables quantitative analysis across diverse imaging modalities and length scales.}
\textbf{a} Unsupervised denoising algorithms predict smoothed data from a partially-masked input source.
Video denoising with \acrfull{umvd} uses a temporal masking scheme \cite{aiyetigbo_unsupervised_2024} to predict individual frames, and image denoising with \acrfull{n2v} uses a spatial masking scheme \cite{krull_noise2void_2019} to predict single images.
Denoising models are trained by minimizing the error between predicted outputs and unmasked noisy inputs.
\textbf{b} Data from a prototypical pattern-forming system is simulated via the Cahn-Hilliard equation and subsequently corrupted with noise and denoised.
Deep denoising is validated using PDE-constrained optimization to recover the ground truth physics from denoised data.
\textbf{c}  At the nanoscale, Lithium iron phosphate nanoparticles are imaged using STXM ~\cite{lim_origin_2016,deng_correlative_2021}.
Raw absorbance measurements denoised with N2V reveal clearer intraparticle heterogeneity in spatiotemporal analysis of lithium composition.
\textbf{d} Mesocale optical microscopy visualizes phase-transformation-induced color changes within a graphite electrode \textit{in-situ}.
Denoising enables large-scale data processing by producing robust segmentation and phase quantification across hundreds of particles.
\textbf{e} At the macroscopic scale, \textit{operando} neutron radiography captures lithium transport across full electrodes within a working cell \cite{gober_high_2025}.
Denoising reduces variability in differential imaging, enabling reliable interpretation of reversible and irreversible transport pathways.
}\label{fig:overview_schematic}
\end{figure}

In this work, we integrate state-of-the-art deep learning denoising algorithms for images and videos into both synthetic and experimental microscopy workflows, and demonstrate their ability to enhance the qualitative and quantitative characterization of heterogeneity in lithium-ion battery materials (Fig. \ref{fig:overview_schematic}).
By improving data quality, our approach reduces uncertainty in model identification with minimal bias, reveals previously obfuscated heterogeneity in scanning transmission X-ray microscopy (STXM) of lithium iron phosphate (LFP), enables automated large-scale analysis in optical microscopy of a graphite electrode, and uncovers spatially heterogeneous lithium transport in neutron radiography of graphite/nickel-manganese-cobalt oxide (NMC) cells.
These results demonstrate that deep learning denoising can be broadly applied across a range of \textit{in situ} microscopy modalities to enhance characterization of heterogeneity and to streamline image-processing pipelines spanning the nanoscale to the device scale.

\subsection*{Unbiased model recovery in simulated data with reduced uncertainty}\label{sec:synth_data}

State-of-the-art microscopy-based material characterization workflows combine mechanistic modeling with image-learning to elucidate and validate constitutive material relationships \cite{kalinin_bigdeepsmart_2015,zhao_learning_2020,deng_correlative_2021,zhao_learning_2023}.
However, measurement noise often obscures signals of interest, introducing uncertainty that hampers the accurate recovery of underlying physical laws \cite{zhao_image_2021}.
Deep learning denoising offers a promising route to overcome this challenge; however, prior studies have highlighted that pixel-level intensity quantification may introduce artifacts and distortions that hinder downstream analysis \cite{laine_imaging_2021}.
Here, using a synthetic dataset, we extend deep denoising validation beyond conventional image quality metrics to directly test how denoising influences the recovery of underlying physics.



\begin{figure}[H]
\centering
\includegraphics[width=\linewidth]{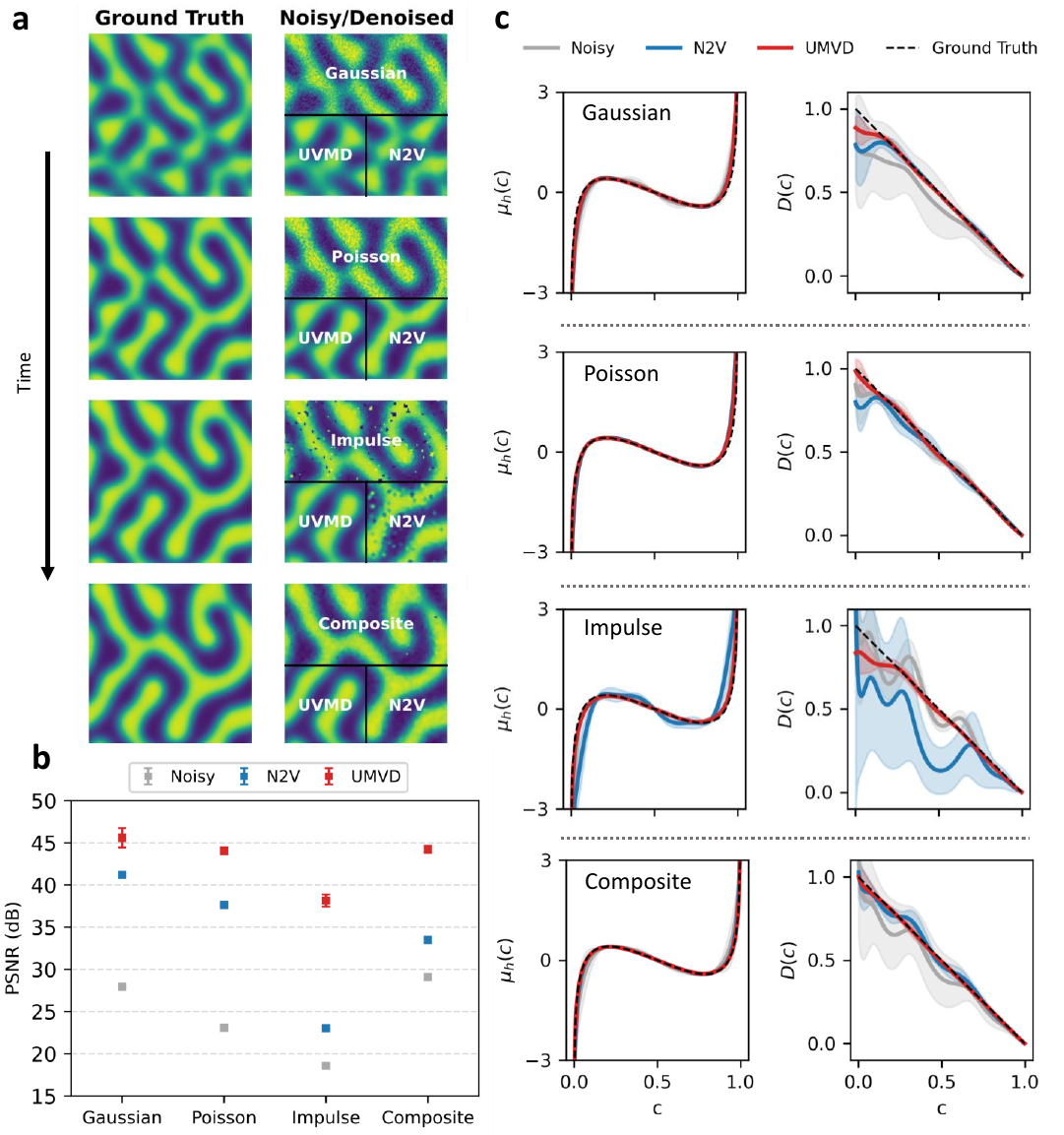}
\caption{\textbf{Denoising workflow increases signal to noise ratio of simulated dataset and recovers the ground truth physical properties with reduced variance.}
\textbf{a} Across three standard noise distributions and a mixed composite case, video denoising (UMVD) achieves visually accurate reconstructions of the ground truth, while image denoising (N2V) provides notable improvements in image quality under continuous noise conditions, such as Gaussian and Poisson.
\textbf{b} UMVD and N2V significantly increase the PSNR of the noisy images, with stronger performance in UMVD.
Similar improvements are observed for the structural similarity index metric (Fig.~\ref{fig:chr_ssim}).  
\textbf{c} Across all noise distributions, UMVD reliably recovers the ground truth $\mu_h(c)$ and diffusivity $D(c)$ with substantially reduced uncertainty, while N2V recovers the ground truth in all but the impulse noise case.
A small discrepancy persisted near $c=0$ for $D(c)$, where insensitivity to $D(c)$ introduced higher uncertainty for all noise conditions \cite{zhao_learning_2020}.
A detailed comparison of final optimization error and results for additional noise intensity levels are provided in Supplementary Information section \ref{subsec:SI_CHR}.
Shaded regions indicate one standard deviation obtained by bootstrapping.
}\label{fig:synth_CHR}
\end{figure}

We illustrate this by corrupting and denoising a synthetic dataset generated by simulating the Cahn–Hilliard equation~\cite{cahn_free_1958,cahn_phase_1965}, a canonical model for pattern-forming systems in materials science~\cite{han_electrochemical_2004,bazant_theory_2013}:
\begin{equation}
\frac{\partial c}{\partial t} = \nabla \cdot \left( D(c)c \nabla \frac{\delta G}{\delta c} \right)\label{eq:CHR_model}
\end{equation}
where $c(x,t)$ is the conserved concentration field, $D(c)$ is a composition-dependent diffusivity, and $\delta G / \delta c$ is the diffusive chemical potential, the variational derivative of the free energy:
\begin{equation}
\frac{\delta G}{\delta c} = \mu_h(c) - \kappa \nabla^2 c
\end{equation}
where $\mu_h(c)$ is the homogeneous chemical potential and $\kappa$ is the gradient penalty parameter that produces diffuse interfaces between low- and high-density phases.
The ground-truth data is corrupted with Gaussian, Poisson, and impulse (salt-and-pepper) noise of varying intensity, as well as a composite model blending these distributions to mimic experimentally relevant conditions (Fig. \ref{fig:synth_CHR}a).
Denoising was performed using both the video-based unsupervised microscopy video denoising (UMVD) algorithm \cite{aiyetigbo_unsupervised_2024} and the image-based Noise2Void (N2V) algorithm \cite{krull_noise2void_2019} (Fig. \ref{fig:synth_CHR}a, \nameref{sec:methods}).
To test whether the denoising model biases the underlying physics, we simulated the Cahn-Hilliard equation with a known diffusivity and chemical potential benchmark (refs. \cite{zhao_learning_2020,zhao_image_2021}) and attempted to recover these functions from a limited number of noisy and denoised snapshots.
The diffusivity and chemical potential were parameterized with orthogonal polynomials, and the coefficients recovered using PDE-constrained optimization (\nameref{sec:methods}, Fig. \ref{fig:synth_CHR}c).

\begin{figure}[H]
\centering
\includegraphics[width=.75\textwidth]{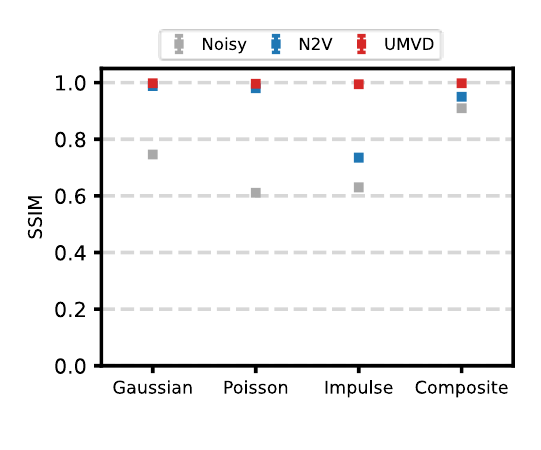}
\caption{\textbf{Structural similarity index metric (SSIM) of noisy and denoised synthetic data}
Video denoising (UMVD) exhibits high SSIM across all noise types indicating strong performance with up to 65\% increase relative to noisy data.
Image denoising (N2V) exhibits strong performance for Gaussian, Poisson, and composite distributions but suffers in the presence of impulse distributions as discussed in \ref{sec:synth_data}.}\label{fig:chr_ssim}
\figtoc{SSIM of noisy and denoised synthetic data}
\end{figure}

Visual inspection reveals that UMVD achieves near-perfect reconstruction across all noise types, whereas N2V performs well for Gaussian and Poisson noise but produces pronounced artifacts under impulse and composite noise (Fig. \ref{fig:synth_CHR}a, Supplementary Video 1).
Quantitatively, peak signal-to-noise ratio (PSNR), a standard log-scale image quality metric \cite{laine_imaging_2021}, increases by 14.8–20.6 dB for UMVD across all noise types and by 17.3–20.3 dB for N2V in the Gaussian and Poisson cases (Fig. \ref{fig:synth_CHR}b).
Similar improvements are observed in the structural similarity index (SSIM), a perceptual metric of structural fidelity \cite{wang_image_2004}, with UMVD yielding values above 0.99 across all distributions and N2V remaining above 0.95 except under impulse noise (Fig. \ref{fig:chr_ssim}).
These results demonstrate that denoising substantially improves image quality under appropriate conditions while also exposing a limitation of image denoising models, their inability to handle large perturbations from impulse noise that drive pixel intensities far outside the data distribution.
Since impulse noise affects pixels independently in time, video-based models exploit temporal context to suppress the noise, whereas image-based methods merely smooth over the corruption, producing pronounced artifacts.

Across all noise distributions, denoising improved the recovery of physical properties compared to noisy data.
Both UMVD and N2V accurately captured the chemical potential, which showed little sensitivity to noise, but their benefit was most evident in recovering the diffusivity, where noisy data exhibited large variability (Fig. \ref{fig:synth_CHR}c).
UMVD consistently matched the ground truth across all distributions, while N2V performed comparably except under strong impulse noise, where pixel-level artifacts degraded performance.
For the experimentally relevant composite distribution, both algorithms produced diffusivity functions in close agreement with the ground truth with low variability, whereas the noisy data showed spurious higher-order contributions and large uncertainty.

By embedding denoised data into PDE-constrained optimization, we demonstrate that deep learning denoising is as an effective pre-processing strategy that improves image quality, preserves the governing physical mechanisms, and substantially reduces uncertainty in model identification with minimal bias across a wide range of noise distributions.
This generalizable framework unlocks new opportunities to recover governing physical laws and material parameters directly from experimental microscopy data once limited by noise.


\subsection*{Enhanced mapping of compositional heterogeneity in nanoscale X-ray microscopy}\label{sec:stxm}

Synchrotron X-ray microscopy enables direct \textit{operando} mapping of chemical and structural dynamics in battery materials at the nanoscale \cite{lin_synchrotron_2017}.
We analyze a dataset combining prior STXM measurements of LFP particles composed of chemically lithiated, biphasic particles imaged ex situ \cite{deng_correlative_2021} and particles imaged in \textit{operando} undergoing electrochemical cycling \cite{lim_origin_2016}.
To provide a proxy for the ground truth, ex-situ biphasic particles are imaged with X-ray spectro-ptychography, a coherent diffraction technique that reconstructs images with spatial resolution beyond the limits of STXM.
Using the $\text{FeL}_3$ edge, X-ray absorbance measurements are converted to two-dimensional profiles of the local lithium composition averaged along the [010] axis to provide direct insight into chemical heterogeneity.
Previous work used image learning to extract models of nonequilibrium thermodynamics, reaction kinetics and surface heterogeneity in this system; however, the inferred physics had non-negligible uncertainty due to measurement noise \cite{zhao_learning_2023}.
Shot noise and spatially correlated noise introduced by raster scanning obscures internal morphological and compositional features \cite{bak_situoperando_2018}  (Fig. \ref{fig:stxm_autocorr}). 
Here, we apply denoising to suppress scan noise and reveal hidden internal heterogeneity, as validated against X-ray spectro-ptychography and tracked during \textit{in situ} electrochemical cycling.

\begin{figure}[h]
\centering
\includegraphics[width=0.9\textwidth]{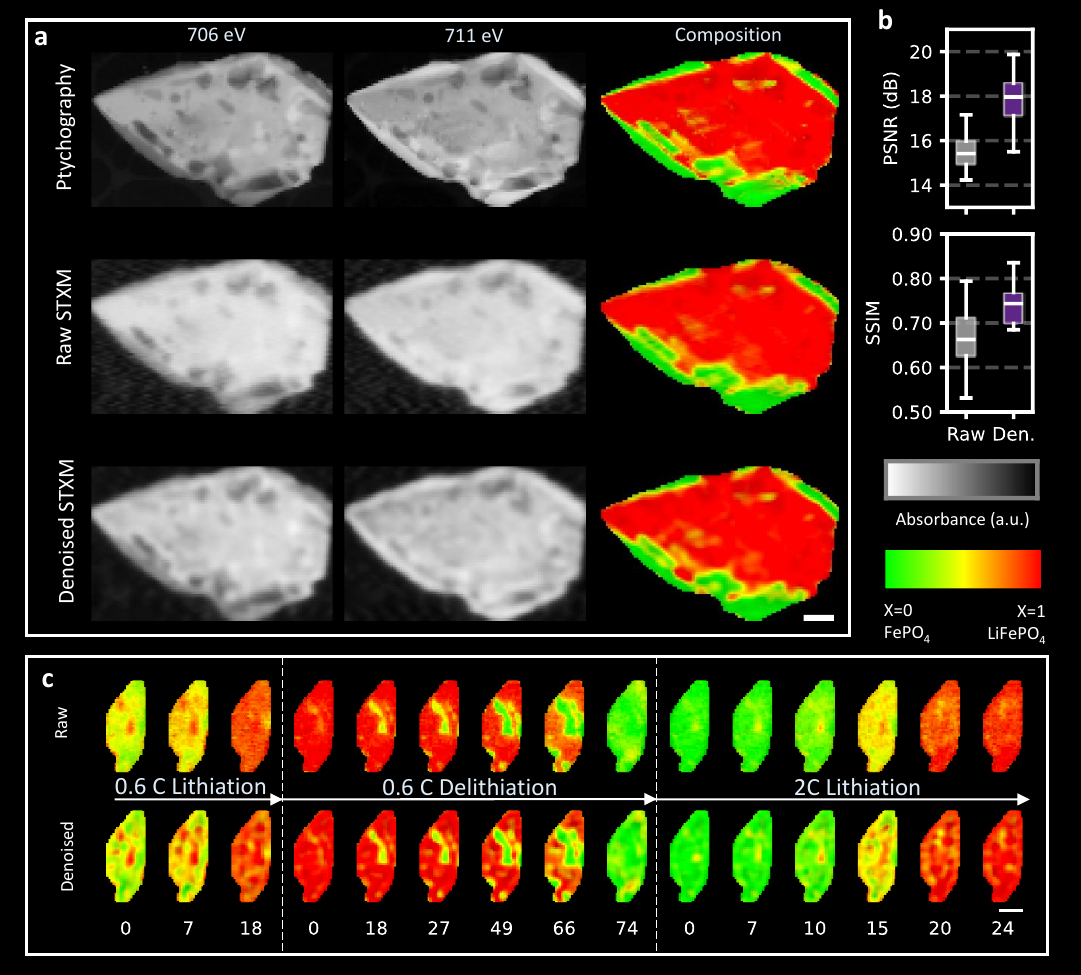}
\caption{
\textbf{Visualization of intraparticle heterogeneity revealed by image denoising in compositional analysis of LFP using STXM.}
\textbf{a} Denoising enhances nanoscale spatial variations in \textit{ex situ} STXM for both X-ray absorbance and lithium composition in biphasic particles, validated against high-resolution X-ray spectro-ptychography.
Noise suppression clarifies intraparticle heterogeneity, bringing STXM measurements into closer agreement with ptychography.
X-ray absorbance measurements were collected at 706eV and 711eV.
Ptychography measurements were resized and aligned to match the STXM images.
\textbf{b} PSNR and SSIM of raw (gray) and denoised (purple) images, computed relative to ptychography and shown as box-and-whisker plots.
Both metrics confirm that denoising improves image quality in terms of both error and perceptual fidelity.
Boxes show the interquartile range (IQR), center lines at the median, and whiskers 1.5$\times$ IQR (n=18 scans, 5 particles).
\textbf{c} \textit{Operando} characterization of LFP after denoising reveals sharper internal phase boundaries and temporally persistent heterogeneity across multiple cycling conditions.
Horizontal scan noise is visually suppressed and phase-separated domains within the denoised particle remain consistent with the raw data and with previous experimental and computational analyses \cite{lim_origin_2016,zhao_learning_2023} 
The C-rate, where $C/n$ stipulates the cell will fully (dis)charge to the theoretical capacity in n hours, is labeled below each half-cycle.
Scale bar: 500 nm; Ex-situ: 40 nm/pixel; In-situ: 50 nm/pixel. 
}\label{fig:lfp_stxm}
\end{figure}

\begin{figure}[H]
\centering
\includegraphics[width=\textwidth]{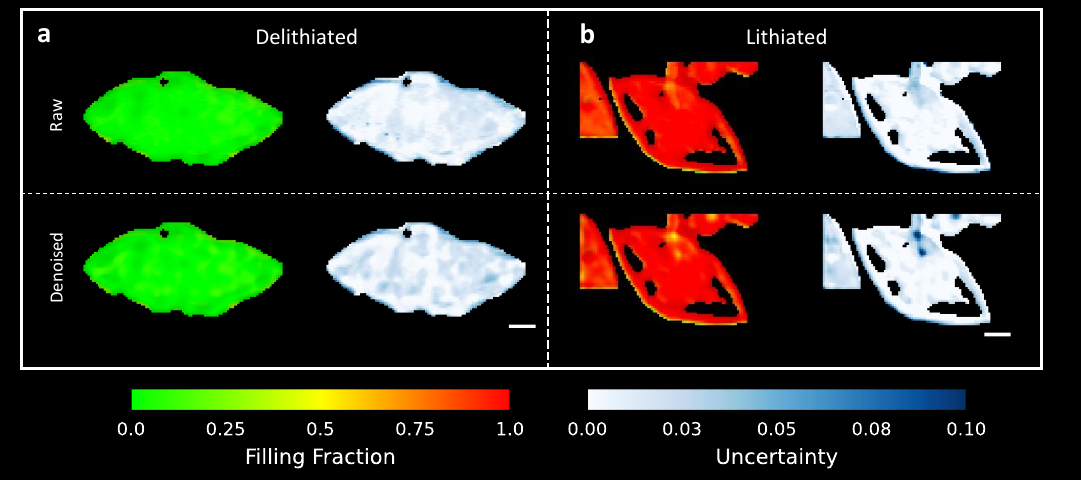}
\caption{\textbf{Lithium composition and uncertainty estimation in reference particles.}  
\textbf{a}, Experimentally mapped lithium composition of a delithiated particle, with corresponding maps of the standard deviation in Li fraction ($\sigma_{\mathrm{Li}}$) obtained by bootstrap resampling for raw and denoised ex-situ images.  
\textbf{b}, Experimentally mapped lithium composition of a lithiated particle, with corresponding $\sigma_{\mathrm{Li}}$ maps from raw and denoised ex-situ images.  
Regions of localized heterogeneity show elevated noise in the composition, reflected as higher variability in both raw and denoised reconstructions. Scale bar 500nm}
\label{fig:stxm_unc}
\figtoc{Composition and uncertainty analysis in reference particles}
\end{figure}

Due to the limited temporal resolution, we apply image denoising using structN2V, a variant of N2V with structured masks that suppresses spatial correlations through rectangular masking during training \cite{broaddus_removing_2020}.
To enhance robustness, we used bootstrap aggregation across 50 independently trained models \cite{breiman_bagging_1996}, yielding consistent performance with an average PSNR of 24.5 dB relative to the raw data.
Processed X-ray absorbance was then converted to lithium filling fraction ($0 < X < 1$) by solving a linear system with $\text{(Li)FePO}_4 $ reference spectra \cite{lim_origin_2016} (\nameref{sec:methods}).
Measurement uncertainty was estimated by bootstrapping the composition calculation to estimate the standard deviation in the inferred lithium fraction \cite{zhao_learning_2023} (Fig. \ref{fig:stxm_unc}).
Across three nearly homogeneous particle images at near-full lithiation or delithiation, the mean deviation in filling fraction was $\sigma_{X}=0.012$ for the raw data and $\sigma_{X}=0.014$ for the denoised data.
The negligible change in uncertainty indicates that denoising has not introduced unphysical artifacts, providing an initial measure of self-consistency that supports its use in subsequent analysis (Fig. \ref{fig:stxm_unc}).

To validate the denoising approach and assess nanoscale heterogeneity, we compared \textit{ex situ} STXM of chemically lithiated, biphasic particles (40 nm pixel size) with high-resolution X-ray ptychography (10 nm pixel size) (Fig. \ref{fig:lfp_stxm}a).
Beyond compositional heterogeneity, these particles exhibit morphological heterogeneity from non-uniform thickness, voids, and heterogeneous strain, as rigorously established by 3D X-ray ptychography–tomography and 4D scanning transmission electron microscopy \cite{deng_correlative_2021}.
Denoising enhanced the visualization of such internal structure in STXM absorbance and composition images, revealing small-scale spatial variations that are otherwise obscured by noise.
Quantitatively, denoising increased the average PSNR of STXM relative to ptychography (used as ground-truth reference) by $15 \%$ across five particles, while the average SSIM rose by $9 \%$ (Fig. \ref{fig:lfp_stxm}b).
Per-pixel correlation analysis shows that denoising strengthens the correspondence between STXM and ptychography absorbance measurements, reducing the root-mean-squared error by approximately 25\% ( Fig. \ref{fig:stxm_ptychocorr}).
Noisy regions in the raw STXM compositional analysis became distinct heterogeneous domains after denoising, particularly in lithiated regions, corresponding directly to the features observed in ptychography (Fig \ref{fig:lfp_stxm}a).
This agreement demonstrates that denoising enhances the interpretability of features at the limit of the instrument’s resolution, revealing fine structural variations previously hidden by noise.

\begin{figure}[H]
\centering
\includegraphics[width=\textwidth]{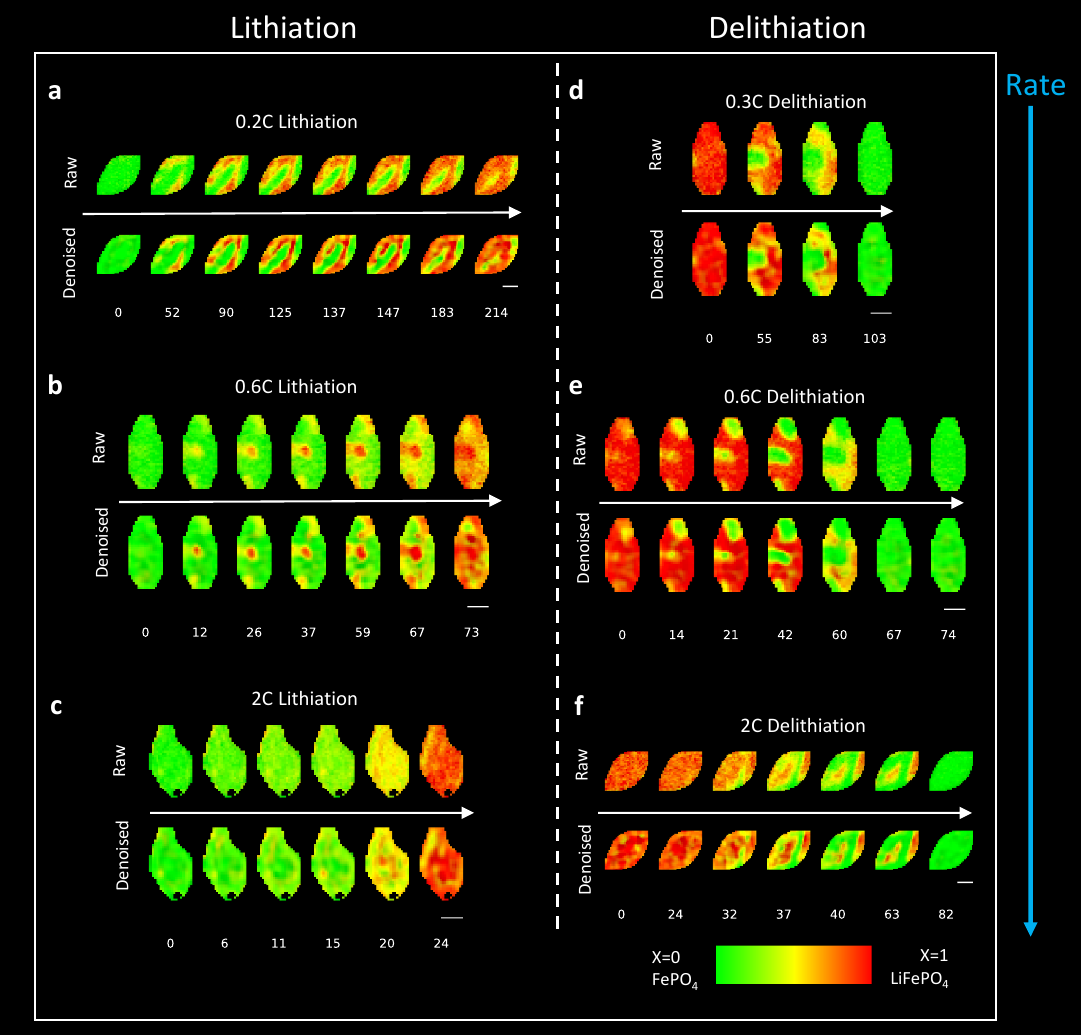}
\caption{\textbf{\textit{Operando} evolution of raw and denoised composition maps for selected particles at increasing rates.}  
\textbf{a}, Low-rate lithiation: pronounced phase boundary formation, consistent with theoretical predictions \cite{lim_origin_2016}.  
\textbf{b}, Intermediate-rate lithiation.  
\textbf{c}, High-rate lithiation: denoising preserves the qualitative signature of auto-inhibition that suppresses phase separation \cite{bai_suppression_2011,bazant_thermodynamic_2017}, as indicated by relatively uniform lithiation. In the final frame, the region of reduced lithiation coincides with an area of lower kinetic activity inferred from pixel-by-pixel inversion \cite{zhao_learning_2023}.  
\textbf{d}, Low-rate delithiation: denoising enhances phase boundaries and improves uniformity within lithiated and delithiated phases.  
\textbf{e}, Intermediate-rate delithiation: denoising renders phase boundaries and propagating intercalation waves more visually distinct.  
\textbf{f}, High-rate delithiation: denoising clarifies the competition between phase separation and reaction kinetics, revealing initial suppression of phase separation followed by stronger separation once the low-lithium phase is nucleated.  
Scale bar 500 nm.}
\label{fig:stxm_extraparticles}
\figtoc{Comparison of dynamics across particles and cycling conditions}
\end{figure}

Across multiple scans and cycling protocols, denoising consistently reduces noise while maintaining the underlying dynamical trends (Fig. \ref{fig:lfp_stxm}c).
Spatial heterogeneity is expected during electrochemical cycling due to the combination of spatially inhomogeneous carbon coating and nonlinear reaction kinetics \cite{lim_origin_2016,zhao_learning_2023}, and denoising sharpens its visualization by clarifying domains that are otherwise obscured by noise.
At higher rates, denoising reveals the uniform lithiation and smooth domains expected from auto-inhibition \cite{bai_suppression_2011,bazant_thermodynamic_2017}.
More broadly, it preserves the expected physical phenomena while enhancing the visualization of heterogeneous dynamics across distinct kinetic regimes, a wide range of cycling rates, and multiple particles (Fig. \ref{fig:lfp_stxm}c, \ref{fig:stxm_extraparticles}).
Notably, the heterogeneous domains revealed by denoising persist across consecutive frames, despite independent processing of each frame, indicating temporal coherence and further validating the methodology.

By improving image quality within the native resolution of the instrument, the denoising approach increases the usable information content of each scan.
In practice, such noise reduction can lessen the need for long-exposure or high-resolution acquisitions, mitigating radiation dose and experimental time while enabling more efficient use of synchrotron beamtime.

\subsection*{Automated, large-scale image processing in mesoscale optical microscopy}\label{sec:optical}

Benchtop \textit{operando} optical microscopy is an accessible and cost-effective tool for studying energy storage materials ~\cite{chen_seeing_2020,maire_colorimetric_2008,harris_direct_2010,merryweather_operando_2021,agrawal_operando_2021}.
In particular, graphite has been extensively studied with color bright-field optical microscopy \cite{maire_colorimetric_2008,harris_direct_2010,agrawal_operando_2021,agrawal_dynamic_2022,guo_li_2016,thomas-alyea_situ_2017} due to its commerical importance \cite{scrosati_lithium_2010} and the well-established correlation between lithium content and color \cite{dresselhaus_intercalation_2002,basu_synthesis_1979}.
Since graphite is a complex, phase-separating material \cite{dahn_phase_1991}, quantifying the multiscale heterogeneity at both the electrode and particle scale is critical to better understand the operational dynamics and failure modes \cite{harris_effects_2013}.
However, most optical studies to date have been limited to single particles \cite{maire_colorimetric_2008,guo_li_2016}, hand-selected subsets of particles \cite{agrawal_operando_2021,agrawal_dynamic_2022,xu_operando_2022}, or bulk electrode averages \cite{thomas-alyea_situ_2017,chen_operando_2021}, rather than providing comprehensive, particle-resolved views of entire electrode regions.
These limitations arise from the difficulty of using qualitative optical signals, which depend strongly on variable illumination conditions and camera response, to resolve complex microstructures and assign RGB values to material phases \cite{agrawal_operando_2021,agrawal_dynamic_2022,guo_li_2016}.
Here, we demonstrate that denoising overcomes these barriers by reducing variability, sharpening particle boundaries, and smoothing the signal distribution.
This enables a streamlined workflow for automated segmentation and phase assignment, extending \textit{operando} optical microscopy from single-particle or bulk case studies to population-level analyses of chemical heterogeneity in graphite electrodes.

\begin{figure}[h]
\centering
\includegraphics[width=\textwidth]{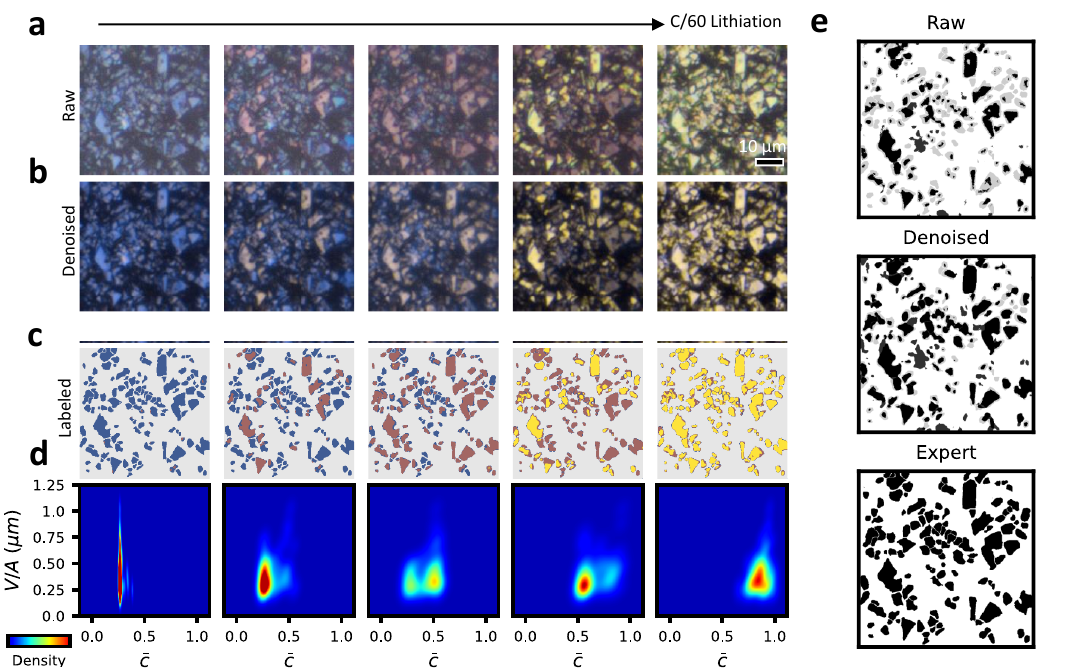}
\caption{\textbf{Video denoising enables electrode-scale particle segmentation and phase classification in optical microscopy of a graphite electrode.}
\textbf{a} Raw frames of graphite during C/60 lithiation experiment.
Graphite undergoes phase transitions from stage III (blue) to stage II (red) to stage I (gold).
Images exhibit cloudiness and short scale temporal fluctuations (Supplementary Video 2).
\textbf{b} Denoised frames enhance particle contrast, reducing haziness and improving particle-background separation. 
\textbf{c} Partic.le segmentations from raw and denoised videos compared with expert manual annotation; denoised results closely match manual expert segmentation.
Overlayed gray regions denote expert segmentation.
\textbf{d} Algorithmic phase classification of masked particles from denoised images with full spatial resolution.
Classifications qualitatively match the corresponding observations in both raw and denoised images.
\textbf{e} Automated denoising workflow enables tracking population-level statistics with particle-level resolution.
Population density is plotted at the corresponding frames as a function of the particle-averaged state of charge, $\bar{c}$, and characteristic particle size $V/A$.
Scale bar: 10 $\mu$m.
 }\label{fig:graphite_optical}
\end{figure}

\begin{figure}[H]
\centering
\includegraphics[width=\textwidth]{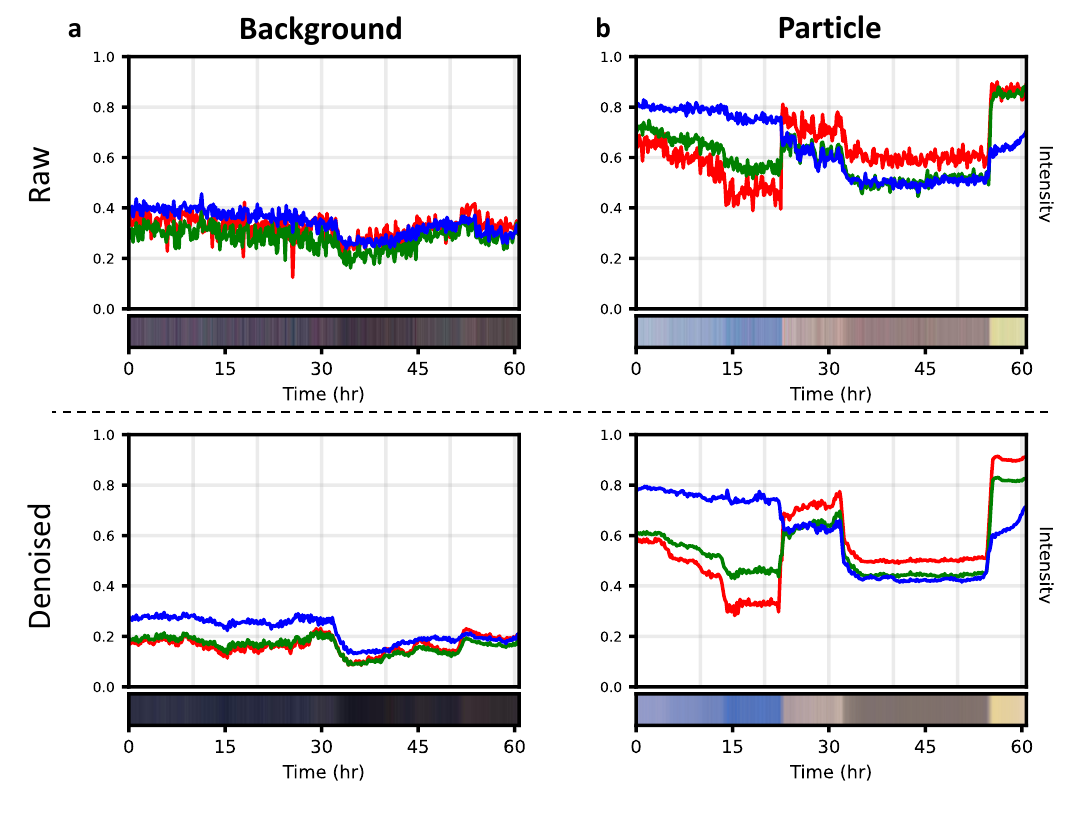}
\caption{\textbf{Time-series RGB traces for representative pixels in the optical microscopy experiment.}  
Colored lines indicate the intensity of the corresponding RGB channels.  
\textbf{a}, A background pixel exhibits strong short-timescale fluctuations. Denoising smooths the signal while preserving the perceived color, as indicated by the color bars.  
\textbf{b}, A pixel within a particle shows similarly smoothed RGB traces after denoising, yielding a more reliable color assignment for each distinct graphite phase.}
\label{fig:om_rgbpixtrace}
\figtoc{Comparison of RGB time-series}
\end{figure}

In this study, a platelet graphite electrode was imaged under slow-rate (C/60) constant-current conditions using \textit{operando} color bright-field microscopy (Fig \ref{fig:graphite_optical}a, Supplementary Video 2).
Raw image sequences exhibited typical shot noise as well as cloudiness and flickering due to variable illumination intensity (Fig. \ref{fig:om_rgbpixtrace}).
To address this, we apply video denoising with UMVD, achieving a validation PSNR of 28.5 and SSIM of 0.728, alongside substantial qualitative improvements.
Denoised images show suppressed variability, enhanced signal fidelity within particles, and clarified color changes (Fig. \ref{fig:graphite_optical}a,b, Supplementary Video 2, Fig \ref{fig:om_rgbpixtrace}).
These enhancements in data quality and contrast between particles and background enabled a fully automated workflow for particle segmentation and pixel-level phase assignment (Fig \ref{fig:graphite_optical}c,d).

To identify a map of all primary particles in the electrode, we performed automated segmentation using $k$-means clustering and benchmarked the results against manual expert segmentation for evaluation (Fig \ref{fig:graphite_optical}c, \nameref{sec:methods}) \cite{guo_li_2016,laine_avoiding_2021}.
Segmentations from the denoised data closely matched the expert annotations, whereas the raw segmentation predominantly identified only the largest particles, occasionally including fragmented regions or artifacts, but failed to capture many small to medium-sized particles.
Lognormal fits to the distributions of characteristic particle size, defined as the volume-to-area ratio assuming constant thickness, highlight this improvement. 
The expert annotation yielded $\mu = 0.40~\mu\text{m}$ and $\sigma = 0.13~\mu\text{m}$; the denoised segmentation reproduced this distribution with $\mu = 0.37~\mu\text{m}$ and $\sigma = 0.15~\mu\text{m}$, while the raw segmentation underestimated both mean size and spread with $\mu = 0.29~\mu\text{m}$ and $\sigma = 0.11~\mu\text{m}$ (Fig.~\ref{fig:om_sizedist}).

\begin{figure}[H]
\centering
\includegraphics[width=\textwidth]{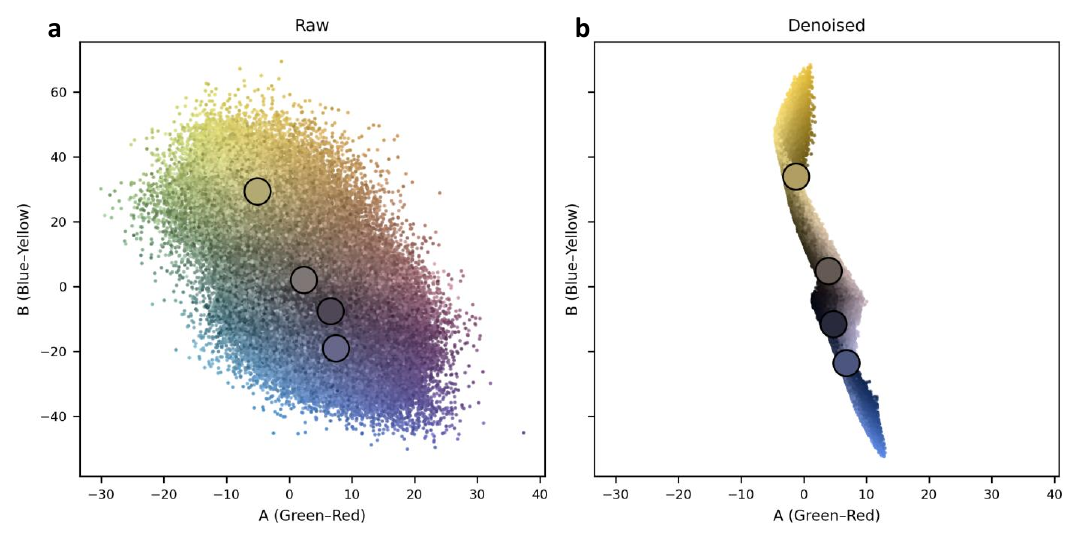}
\caption{\textbf{Optical microscopy pixel distributions in the AB coordinates of LAB color space with $k$-means clustering ($k=4$).}  
\textbf{a}, Raw data show a broad distribution of pixel values without sharp density around the expected colors: black/dark purple (background), blue (stage III graphite), red–brown (stage II graphite), and gold (stage I graphite).  
\textbf{b}, Denoising sharpens the distribution, confining it to a clearer manifold with distinct cluster centers corresponding to the expected graphite phases and background.  
For visualization, only the AB coordinates are shown, omitting the L channel.}
\label{fig:om_datadist}
\figtoc{Optical microscopy RGB data distributions}
\end{figure}

After applying the segmentation mask, we predicted a spatiotemporally resolved map of the phases from the denoised images using a color-clustering algorithm (Fig.~\ref{fig:graphite_optical}d, \nameref{sec:methods}).
Denoising improved performance by smoothing the data distribution, (Fig. \ref{fig:om_datadist}) enabling more stable clustering.
Furthermore,
it suppressed short-scale fluctuations in phase assignments caused by noise, yielding robust intraparticle signals (Supplementary Video 2).
The resulting classifications were consistent with the expected phase behavior observed in the raw  microscopy data (Fig \ref{fig:graphite_optical}a,d , Supplementary Video 2).
To the author's knowledge, this workflow achieves the first large-scale ($N > 100$) quantification of heterogeneous particle dynamics within a graphite electrode.
The curated dataset provides sufficient statistics for population-level analysis, enabling tracking of density distributions as a function of time, average filling fraction, and particle size (Fig. \ref{fig:graphite_optical}e).
At slow rates, the population exhibits characteristic phase-separating dynamics, mainly bimodal distributions within the miscibility gap \cite{zhao_population_2019}.
Particle-level resolution further reveals size effects: the stage III - II transition proceeds stochastically and is largely independent of particle size, while the stage II-I transition shows clear size dependence, with two phases of concurrent intercalation waves, first among small particles, then among large ones.

By improving data quality across the full field of view, denoising removes the need to restrict analysis to hand-selected regions, reducing observer bias and enabling statistically representative characterization across hundreds of particles.
More broadly, the ability to analyze entire electrodes rather than small, curated patches increases throughput and reproducibility, transforming \textit{operando} optical microscopy from a qualitative observation into a quantitative, population-level measurement.


\subsection*{Revealing cell-scale transport pathways in neutron radiography}\label{sec:neutron}

\begin{figure}[H]
\centering
\includegraphics[width=.89\textwidth]{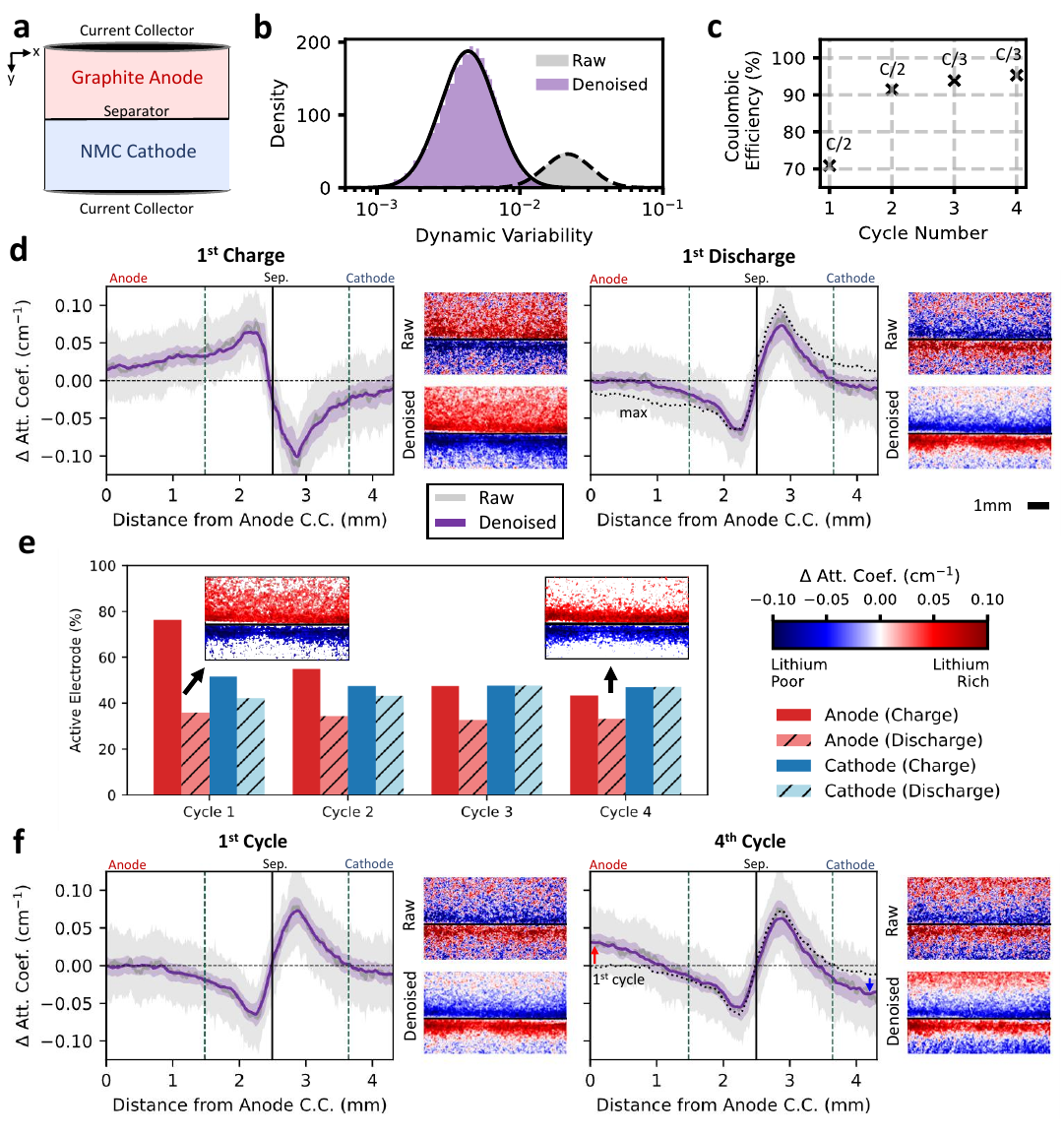}
\caption{\textbf{Reduced variability enables mechanistic insight into macroscopic heterogeneity in \textit{operando} neutron radiography.}
\textbf{a} Schematic of the analyzed graphite-NMC test cell with thick electrodes and a solvent-free anode preparation.
\textbf{b} Distribution of dynamic variability, defined as the local spatiotemporal standard deviation (3 × 3 × 3 window).
Denoising lowers the mean from 0.0247 (raw) to 0.005 (denoised), nearly an 80\% reduction.
\textbf{c} Coulombic efficiency versus cycle number with C-rates indicated.
The first cycle displays markedly lower efficiency, before stabilizing in later cycles.
\textbf{d} Change in attenuation coefficient reveals uniform anode filling during charge, indicated by the flat distribution across the anode, whereas discharge is heterogeneous, indicated by sharp localized peaks confined to a small active region (dashed green lines).
Full 2D maps and corresponding 1D depth averages are shown: the raw data (gray) exhibits high variability with broad error bars, whereas the denoised data (purple) suppresses fluctuations and highlights non-uniform discharge that leaves a zone of inactive lithium near the current collector.
Shaded regions denote one standard deviation in the in-plane (x-axis) attenuation. 
\textbf{e} Active area fraction for both electrodes across half-cycles.
The first charge cycle shows high anode activity, whereas later cycles exhibit reduced and asymmetric utilization between charge and discharge, consistent with lithium sequestration. 
By the fourth cycle, activity is confined to a localized region near the separator, as shown in the inscribed visualization.
\textbf{f} Change in lithium attenuation relative to the end of the first charge cycle, shown for both raw and denoised radiographs with corresponding 1D averages.
Denoising reduces variability and reveals a clear progression of lithium sequestration near the current collector, with a smooth gradient across the anode.
Scale bar: 1 mm.
}\label{fig:neutron}
\end{figure}

With the emergence of advanced high-resolution instruments, neutron imaging is an increasingly promising technique for probing transport phenomena and degradation processes in lithium-ion batteries \cite{owejan_direct_2012,siegel_neutron_2011,ziesche_neutron_2022}.
We analyze an \textit{operando} dataset of a NMC-graphite full cell undergoing electrochemical cycling with a constant current constant voltage (CC-CV) protocol \cite{gober_high_2025}.
The anode is prepared with a solvent free preparation, leading to nontrivial, heterogeneous transport in the depth direction of the battery (Figure \ref{fig:neutron}a).
To track lithium transport, the data analysis requires normalizing the raw intensity to a reference frame (\nameref{sec:methods}); however, the division of noisy frames severely amplifies variability, obscuring meaningful signals and hampering quantitative analysis.
We show that applying deep learning-based denoising substantially reduces this variability in differential neutron imaging, unlocking new opportunities for the visualization and quantification of dynamic mesoscale transport processes in lithium-ion batteries.
This workflow allows us to mechanistically probe cell degradation at the macroscopic electrode level and reliably demonstrate that the solvent-free microstructure leads to lithium sequestration near the anode current collector.

We first apply denoising to the raw neutron transmission data, achieving a validation PSNR of 35.7 dB and SSIM of 0.90 relative to the raw data.
Intensities were normalized and converted to the attenuation coefficient to evaluate changes in lithium transport:
\begin{equation}
    \Delta \Sigma_{Li}(t) = \Sigma(t) - \Sigma(t_{ref}) = \frac{1}{\delta}\log \left(\frac{T(t)}{T(t_{ref})}\right) \label{eq:att_norm}
\end{equation}
where $\Delta \Sigma_{Li}$ is the change in lithium attenuation coefficient, $t_{ref}$ the reference time point, $\delta$ the sample thickness, and $T$ the normalized neutron transmission (Supplementary Video 3, \nameref{sec:methods}).
Denoising reduced dataset variability by nearly an order of magnitude, yielding spatiotemporal smoothness consistent with the intrinsic resolution of the measurement (3 pixels), below which single-pixel variations are not physically meaningful \cite{gober_high_2025}.
With this improvement, $\Delta \Sigma_{\text{Li}}$ maps provide a more reliable view of dynamic lithium transport.
A notable feature in this dataset is the pronounced capacity loss during the first cycle (Fig. \ref{fig:neutron}c) where the Coulombic effiency is only 70\%, indicating a 30\% loss of the initial capacity.
Identifying where this lithium is sequestered and how it evolves with subsequent cycling is crucial to diagnosing the degradation mechanisms in this system.

\begin{figure}[H]
\centering
\includegraphics[width=\textwidth]{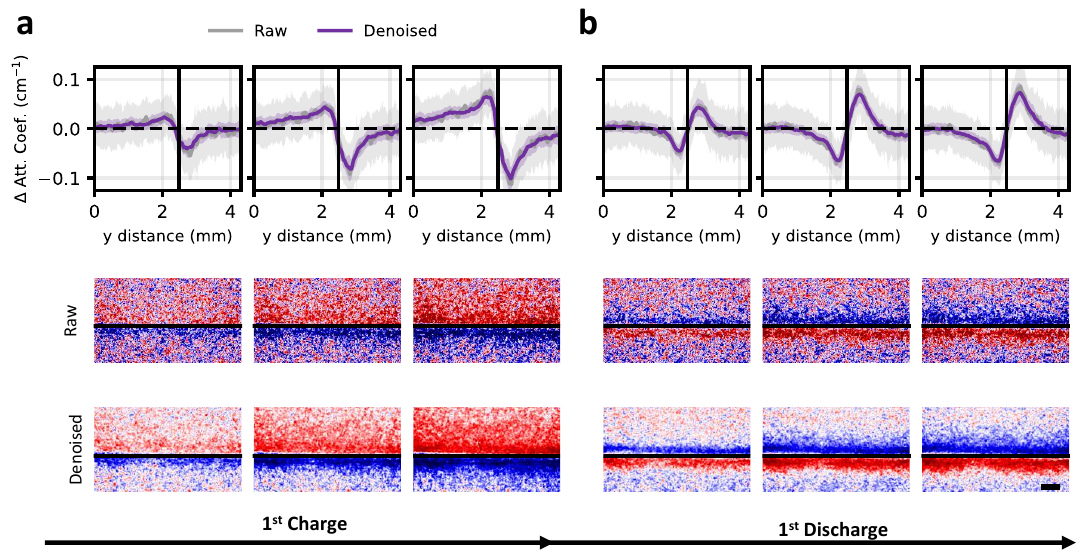}
\caption{\textbf{\textit{Operando} tracking of attenuation changes during the first electrochemical cycle.}  
\textbf{a}, During the first charge, the anode fills relatively uniformly, while the cathode depletes more sharply near the separator. 
Denoising reveals a smooth spatial gradient, as shown in the one-dimensional depth-averaged profiles.  
\textbf{b}, During the first discharge, transport is heterogeneous, with activity confined to the regions near the separator in both the anode and cathode.  
The denoised image sequences highlight a smoother spatiotemporal progression of attenuation over time.}
\label{fig:nr_firstcycle}
\figtoc{Neutron radiography of asymmetric dynamics in first cycle}
\end{figure}

To investigate the origin of the low Coloumbic effiency in the first cycle, we visualized lithium dynamics relative to the onset of the first charge and discharge half-cycles using  raw and denoised radiographs together with depth-averaged profiles (Fig \ref{fig:neutron}d, \ref{fig:nr_firstcycle}).
While raw and denoised averages appear similar, the uncertainty, quantified as $\pm 1$ standard deviation along the depth direction, is substantially higher for the raw data.
In the full 2D radiographs, this arises from short-scale spatial fluctuations amplified during temporal normalization given by Eq. \ref{eq:att_norm}.
Denoising suppresses these fluctuations, yielding smoother signals and enabling more reliable quantitative analysis.
With reduced variability, the denoised data reveals a homogeneous anode filling during the first charge cycle, culminating in a uniformly lithiated state.
In contrast, the first discharge is highly non-uniform: the half of the anode adjacent to the current collector fails to delithiate (Fig \ref{fig:neutron}d, Supplementary Video 3).
In the raw data, this region shows spurious increases in lithium attenuation, whereas the denoised data clearly resolves a smooth spatial gradient that indicates a lack of delithiation.
These findings suggest that the capacity lost between the first charge and discharge remains sequestered in this inactive region.

\begin{figure}[H]
\centering
\includegraphics[width=\textwidth]{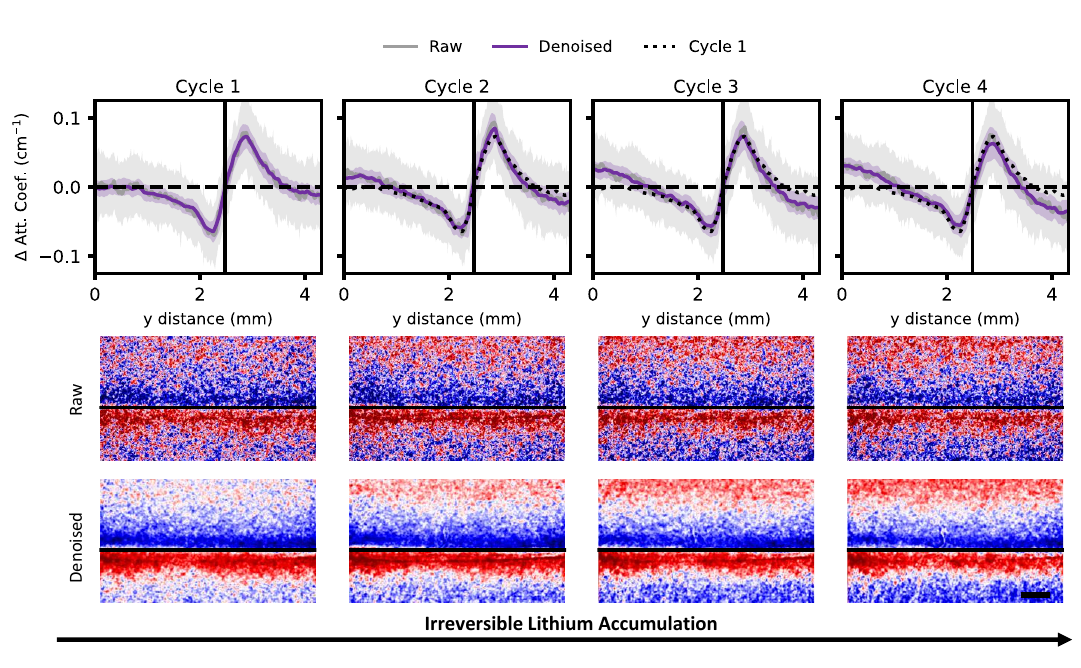}
\caption{\textbf{Lithium sequestration near the anode current collector over multiple cycles.}  
One-dimensional depth-averaged attenuation, referenced to the first discharged state, reveals lithium accumulation near the anode current collector and corresponding depletion near the cathode side.  
Denoising highlights a smooth gradient across the anode, providing strong evidence of lithium trapping: lithium progressively accumulates with each charge cycle and does not fully deplete on subsequent discharges.}
\label{fig:nr_lithacc}
\figtoc{Accumulation of inactive lithium during cycling}
\end{figure}

Distinct attenuation distributions in the denoised data enabled thresholding of both electrodes to identify active regions during cycling (Fig. \ref{fig:neutron}e).
In the first charge, nearly 80\% of the anode is active during charge compared to only 50\% of the cathode.
Upon discharge, anode activity dropped below 40\%, while the cathode remained at approximately $45$\% active.
This pronounced asymmetry supports a mechanism in which lithium becomes sequestered in inactive regions near the current collector, reducing effective capacity.
Extending the analysis across subsequent cycles revealed persistent anode asymmetry, suggesting progressive lithium accumulation, while the cathode remained largely symmetric between charge and discharge.
To visualize this accumulation, we computed the change in attenuation in the discharged state relative to the first charge half-cycle (Fig \ref{fig:neutron}f, \ref{fig:nr_lithacc}).
This approach revealed substantial lithium growth in the anode near the current collector.
Denoising again reduced uncertainty and uncovered a clear spatial gradient with a region of expected depletion near the separator, a band of unchanged attenuation in the mid-electrode, and pronounced lithium accumulation adjacent to the current collector.
These results indicate that charge–discharge asymmetry in the anode drives significant lithium trapping near the current collector, consistent with previous experimental findings \cite{gober_high_2025}.
Denoised \textit{operando} neutron radiography provides mechanistic insight through reliable one- and two-dimensional visualizations that are otherwise obscured by variability in the raw data.
More broadly, the denoising framework enhances both the fidelity and efficiency of neutron imaging, extending its usefulness for operando studies where experimental time is inherently limited.

\section*{Conclusion and Outlook}\label{sec:conclusion}
We present a framework for integrating unsupervised deep denoising into quantitative \textit{operando} materials microscopy workflows.
Two state-of-the-art algorithms, UMVD and N2V, are rigorously validated for model-based characterization, showing that denoising preserves the fidelity of underlying physics recovered via PDE-constrained optimization.
As a preprocessing step, denoising enhances X-ray microscopy pipelines by improving the visualization of nanoscale heterogeneity in LFP, as confirmed by high-resolution ptychography and dynamical analysis.
In optical microscopy, it enables automated, electrode-scale particle segmentation and phase classification for graphite.
For differential imaging, such as \textit{operando} neutron radiography, denoising proves essential for reliably resolving transport and degradation processes.
Together, these findings highlight deep denoising as a versatile and powerful tool for advancing quantitative analysis across a broad range of microscopy modalities, length scales, and workflows.
This framework addresses two persistent challenges in the field: mitigating the trade-offs between resolution, signal quality, and noise that limit beamline efficiency, and enabling high-throughput, pixel-level quantitative analysis in benchtop microscopy.
To facilitate community adoption, we have released an open-source Python package for video denoising, enabling microscopy practitioners to readily integrate these methods into existing experimental and analysis workflows.

Looking ahead, this framework opens new directions for joint experimental–computational strategies in the design of future materials microscopy studies.
The demonstrated improvements in operando neutron imaging highlight how deep denoising can elevate data quality in other modalities that face similar limitations.
Techniques such as optical interferometric scattering microscopy (iSCAT) \cite{merryweather_operando_2021} and fluorescence microscopy \cite{fuladpanjeh-hojaghan_-operando_2019}, which depend on ratiometric analysis or reference-frame normalization, are likely to benefit through reduced variability and enhanced signal-to-noise ratio, leading to greater quantitative reliability and robustness.
Atomic force microscopy, often constrained by high noise levels \cite{joo_atomic_nodate}, stands to benefit substantially from enhanced signal fidelity and noise suppression.
More broadly, advances in unsupervised machine learning for image preprocessing open new opportunities for quantitative \textit{in situ} studies, enabling the use of previously inaccessible modalities and supporting data-driven discovery of constitutive physical laws.

\section*{Methods}\phantomsection\label{sec:methods}
 
\subsection*{Video and Image Denoising}
\phantomsection
\label{subsec:denoise_meth}
Video denoising networks were trained using a spatiotemporal U-Net architecture \cite{aiyetigbo_unsupervised_2024} in a distributed data-parallel setting on two NVIDIA V100 GPUs.
Models were optimized with Adam or AdamW, batch size of 1, and run for up to 50 epochs or until the cluster job time limit of 96 hours was reached.
Dataset-specific training parameters, including input size, spatial patch dimensions, number of temporal frames, and learning rate, are summarized in Table~\ref{tab:denoise_datasets}.
Image denoising was performed using the self-supervised Noise2Void framework \cite{krull_noise2void_2019} with a U-Net backbone implemented in Careamics.
For the simulated dataset, the N2V2 variant \cite{hock_n2v2_2023} was applied with random flips and rotations as data augmentation.
The LFP STXM data was denoised using structN2V \cite{broaddus_removing_2020} with a horizontal mask size of 7 and data augmentation by random flips.
In all cases, images were sub-sampled into patches and trained with the Adam optimizer, using the dataset-specific parameters summarized in Table~\ref{tab:image_denoise_datasets}.
For robustness, 50 models were trained per dataset using ensemble bootstrapping, and final predictions were obtained by averaging across ensemble outputs.

\subsection*{Cahn-Hilliard Simulations and PDE-constrained Optimization}\phantomsection\label{subsec:CHR_sim_meth}

Synthetic training data was generated by simulating the Cahn–Hilliard equation (Eq.~\ref{eq:CHR_model}) with a concentration-dependent diffusivity and regular solution chemical potential under periodic boundary conditions.
Ten realizations were produced on $128 \times 128$ grids, spanning early spinodal decomposition to late-stage coarsening.
To evaluate denoising robustness, ground-truth images were corrupted with noise from the following distributions: Gaussian, Poisson, Impulse, and a composite noise model at varying intensities.
The unknown diffusivity and chemical potential functions were parameterized using Legendre polynomials and inferred using a PDE-constrained optimization procedure based on multiple shooting.
The simulations and optimizations were performed using mosaix-pde \cite{cohen_mosaix-pde_2025}, which leverages JAX for GPU acceleration and automatic differentiation and diffrax \cite{kidger_neural_2022} for solving the discretized PDEs.
Optimization employed a Levenberg–Marquardt algorithm with \(L_2\) regularization.
Full details of the simulation parameters, noise levels, optimization settings, and numerical solver configurations are provided in Supplementary Information section \ref{subsec:SI_CHR}.

\subsection*{STXM Image Processing}\phantomsection\label{subsec:STXM_meth}
Raw STXM datasets from previous \textit{ex situ} \cite{deng_correlative_2021} and \textit{in situ} \cite{lim_origin_2016} studies were reprocessed following the denoising procedure described in Subsection ~\ref{subsec:denoise_meth}.
Absorbance images were cropped to particle boundaries, registered using phase cross-correlation \cite{guizar-sicairos_efficient_2008}, and converted to optical density, which was mapped to lithium fraction using reference Fe–$L_3$ edge spectra.
For \textit{in situ} data, lithium fraction was determined by solving a two-energy linear system after pre-edge normalization, whereas for \textit{ex situ} data, non-negative least squares fitting was applied across all energy levels.
Uncertainty was estimated by bootstrapping across energies, yielding $\sigma_X = 0.012$–$0.014$ with $n=65$ levels, an 83\% reduction relative to previous estimates \cite{zhao_learning_2023}.
Ptychography maps were processed analogously, resized, and spatially aligned to STXM for direct comparison.
Full details of composition determination, uncertainty analysis, and registration procedures are provided in the Supplementary Information section \ref{subsec:SI_STXM}.

\subsection*{Graphite Optical Microscopy Experiment}\phantomsection\label{subsec:OM_exp}
Graphite flakes (7--10~µm diameter, Alfa Aesar), copper foil (9~µm, MSE Supplies), conductive acetylene black, HSV900 PVDF binder, and 1-methyl-2-pyrrolidone (NMP) were obtained from MSE Supplies.
Celgard ceramic-coated separators (Q18G1SY) and Whatman glass fiber separators were used for \textit{operando} cell assembly. The electrolyte was 1~M LiPF\textsubscript{6} in EC/DMC (50:50~v/v, MSE Supplies).
A slurry of graphite, acetylene black, and PVDF (90:5:5 by weight) was prepared in NMP and cast onto the ceramic-coated separator.
The coated separator was dried under vacuum at 60~\textdegree C for 3~days to remove NMP and moisture, yielding an areal loading of 0.8--1.0~mg~cm\textsuperscript{$-2$}.
Operando CR2032 coin cells were assembled by drilling a hole in the positive case and sealing a thin glass window with epoxy.
A copper foil (1 mm central hole) was placed beneath the window, followed by the graphite-coated separator, a glass fiber soaked with electrolyte, and a lithium metal counter electrode.
The cell was completed with a spacer and spring and crimped in an argon-filled glovebox.
Cells were imaged under an Olympus B53M microscope with top illumination to observe graphite particles during cycling.
Formation was performed at C/5 for three cycles prior to measurements.

\subsection*{Optical Microscopy Image Processing}\phantomsection\label{subsec:OM_meth}

The optical microscopy video of a C/60 lithiation–delithiation cycle was denoised using UMVD \cite{aiyetigbo_unsupervised_2024} and aligned by drift correction in Nanopyx \cite{saraiva_nanopyx_2023}.
A representative $512 \times 512$ spatiotemporal patch was selected for analysis (Supplementary Fig. \ref{fig:om_patch}). Particles were segmented using k-means clustering in LAB color space, followed by manual expert refinement in Napari \cite{sofroniew_napari_2025} to delineate grain boundaries.
Phase classification was performed on raw and denoised data using clustering with heuristic corrections to enforce physically consistent phase transitions.
Particle size distributions were extracted from labeled objects and fitted with log-normal functions (Supplementary Fig. \ref{fig:om_sizedist}), and population densities were computed by mapping average particle concentration against characteristic size using kernel density estimation (Fig. \ref{fig:graphite_optical}).
To validate the workflow, we estimated image-based state of charge (SOC) by weighting the area fractions of stage III (blue), stage II (red), and stage I (gold) domains by their characteristic concentrations \cite{agrawal_dynamic_2022}, and compared the result to electrochemical SOC, observing close agreement (Supplementary Fig. \ref{fig:om_validation}).
Full segmentation workflows, heuristic rules, and size distribution analyses are provided in the Supplementary Information section \ref{subsec:SI_OM}.

\subsection*{Neutron Radiography Data Processing}\label{subsec:NR_meth}
Raw \textit{operando} neutron radiographs were reprocessed from previously published experiments \cite{gober_high_2025}.
A $3 \times 3$ median filter was applied to the intensity images and to the corresponding open-beam and dark-current references, followed by denoising using the procedure described in Subsection~\ref{subsec:denoise_meth}.
Normalized transmission was calculated from the filtered images and converted to attenuation coefficients using the Beer–Lambert relation, with electrode thickness determined from a cylindrical geometry model and a calibrated pixel-to-distance conversion of 1289.8 pixels per cm.
The field of view was cropped to the solvent-free cell.
For dynamic analysis, the attenuation data were divided into eight half-cycles (four charge, four discharge) and normalized to the first frame of each half-cycle.
At the final frame of each half-cycle, Otsu thresholding \cite{otsu_threshold_1979} was applied to classify pixels as active or inactive, enabling quantification of the fraction of active electrode area (Fig.~\ref{fig:neutron}).
The boundaries of the active window (green vertical lines in Fig.~\ref{fig:neutron}) were defined by the $y$-positions where the active pixel fraction dropped below 20\%, allowing quantitative comparison of electrode utilization across cycles.
Further details on data processing are provided in Supplementary Information section \ref{subsec:NR_meth}.

\section*{Code Availability}\label{sec:code_avail}
The code used for video denoising is available at \url{https://github.com/stmorgenstern/mumvd}.
Additional code for image denoising and data analysis is available from the corresponding authors upon reasonable request.

\section*{Data Availability}\label{sec:code_avail}
The simulated data associated with this paper can be found at \url{https://doi.org/10.6084/m9.figshare.30471311.v1}.
Additional data is available from the corresponding authors upon reasonable request.

\section*{Contributions}\label{sec:credit_authorship}
S.D.M. obtained the first preliminary results and initiated the project.
S.D.M. and M.Z.B. conceived the full study.
S.D.M. performed all denoising and data processing for both simulated and experimental microscopy datasets and developed the video denoising software.
A.E.C. developed and conducted the Cahn–Hilliard simulation and optimization workflows.
R.G. performed the optical microscopy experiments.
M.G. carried out the neutron radiography experiments and assisted in data processing.
G.J.N. supervised the neutron radiography experiments and data processing.
P.B. supervised the optical microscopy experiments and data processing.
M.Z.B. supervised the overall project.
S.D.M. prepared the manuscript.
All authors discussed the results and contributed to the manuscript revision.


\section*{Declaration of Competing Interest}\label{sec:dec}
The authors declare no competing interests.

\section*{Acknowledgements}\label{sec:acknowledgements}
This work was supported by the Toyota Research Institute through D3BATT: Center for Data-Driven-Design of Li-ion Batteries.
S.D.M. acknowledges support from the National Science Foundation Graduate Research Fellowship under Grant No. 2141064.
R.G. and P.B. acknowledge support from the National Science Foundation under Award No. 2044932.
M.G. and G.J.N. acknowledge support from the Shull Wollan Center at the University of Tennessee, the National Science Foundation through Award CBET-1454437, and the DOE Office of Science through Award DE-SC0025220.
This research used resources at the High Flux Isotope Reactor, a DOE Office of Science User Facility operated by Oak Ridge National Laboratory.
The beam time was allocated to MARS under proposal number IPTS-31601.1.
The authors gratefully acknowledge the ORNL Neutron Sciences Team, including James R. Torres, Yuxuan Zhang, Jean-Christophe Bilheux, Hassina Z. Bilheux, and Shimin Tang, for their support.
The authors are grateful to Haitao D. Deng for providing reference STXM analysis code and supporting ptychography data from refs. \cite{deng_correlative_2021,zhao_learning_2023}.
The authors acknowledge the MIT Office of Research Computing and Data for providing high performance computing resources that have contributed to the research results reported within this paper.
The authors also thank Yash Samantaray, Michael L. Li, Amelia Dai, Shakul Pathak, and Brian Carrick for valuable discussions.



\putbib                         
\end{bibunit}

\clearpage

\beginsupplement
\begin{center}
  \LARGE \textbf{Supplementary Information}\\[0.9\baselineskip]
  \large Deep-learning denoising unlocks quantitative insights in \textit{operando} materials microscopy
\end{center}
\vspace{1.2\baselineskip}

\begin{bibunit}[sn-nature]                 











\affil*[1]{\orgdiv{Department of Chemical Engineering}, \orgname{Massachusetts Institute of Technology}, \orgaddress{\street{25 Ames St.}, \city{Cambridge}, \postcode{02139}, \state{MA}, \country{USA}}}

\affil[2]{\orgdiv{Department of Energy, Environmental and Chemical Engineering}, \orgname{Washington University in St. Louis}, \orgaddress{\street{1 Brookings Dr}, \city{St. Louis}, \state{MO}, \postcode{63130}, \country{USA}}}

\affil[3]{\orgdiv{Department of Mechanical and Aerospace Engineering}, \orgname{University of Alabama in Huntsville}, \orgaddress{\city{Huntsville}, \state{AL}, \country{USA}}}

\affil[4]{\orgdiv{Neutron Scattering Division}, \orgname{Oak Ridge National Laboratory}, \orgaddress{\city{Oak Ridge}, \state{TN}, \country{USA}}}

\affil[5]{\orgdiv{Institute of Materials Science and Engineering}, \orgname{Washington University in St. Louis}, \orgaddress{\street{1 Brookings Dr}, \city{St. Louis}, \state{MO}, \postcode{63130}, \country{USA}}}

\affil[6]{\orgdiv{Department of Mathematics}, \orgname{Massachusetts Institute of Technology}, \orgaddress{\street{182 Memorial Dr.}, \city{Cambridge}, \postcode{02139}, \state{MA}, \country{USA}}}












\tableofcontents
\newpage

\section{Supplementary Figures and Tables}\label{sec:SI_figtab}

\subsection{Video and Image Denoising}\label{subsec:SI_figtab_denoise}

\begin{table}[h!]
\centering
\caption{\textbf{Datasets and training parameters for video denoising.} 
Epoch counts marked with an asterisk (*) indicate runs that exceeded the 96-hour cluster job time limit.}
\tabtoc{Datasets and training parameters for video denoising}
\label{tab:denoise_datasets}
\begin{tabularx}{\textwidth}{lXXXX}
\hline
\textbf{Experiment} & \textbf{Dataset Size} ($ T \times H \times W \times C$) & \textbf{Patch Size} & \textbf{Learning Rate} & \textbf{Epochs} \\
\hline
CHR & $501 \times 128 \times 128 \times 1$ & $7 \times 128 \times 128$ & $10^{-4}$ & 50 \\
Graphite & $725 \times 1920 \times 3$; $2560$ & $5 \times 256 \times 256$ & $10^{-3}$ & $2^*$ \\
Neutron Intensity & $502 \times 990 \times 5200 \times 1$ & $7 \times 128 \times 128$ & $10^{-3}$ & $38^*$ \\
Neutron Open Beam \& Dark Current & $5 \times 990 \times 5200 \times 1$ & $5 \times 128 \times 128$ & $10^{-4}$ & 10 \\
\hline
\end{tabularx}
\end{table}

\begin{table}[h!]
\centering
\caption{\textbf{Datasets and training parameters for image denoising.} 
All models used a U-Net backbone trained with self-supervised Noise2Void variants.}
\tabtoc{Datasets and training parameters for image denoising}
\label{tab:image_denoise_datasets}
\begin{tabular}{lccc}
\hline
\textbf{Experiment} & \textbf{Patch Size} & \textbf{Learning Rate} & \textbf{Epochs} \\
\hline
CHR (simulated) & $64 \times 64$ & $10^{-3}$ & 200 \\
LFP STXM & $16 \times 16$ & $5\times10^{-4}$ & 100 \\
\hline
\end{tabular}
\end{table}

\subsection{Simulated Pattern Formation}\label{subsec:SI_figtab_chr}
\begin{table}[h!]
\centering
\caption{\textbf{Results of optimization with Gaussian noise.} Average final mean squared error (MSE) ($\pm$ 1 standard deviation) across ten synthetic dataset realizations for noisy, N2V, and UMVD data, comparing runs with and without physical priors.
The ground truth dataset is corrupted and denoised under 3 increasing levels of relative noise intensity.}
\tabtoc{Results of optimization with Gaussian noise}
\label{tab:gaussian_mse_split}
\small
\begin{tabularx}{\textwidth}{lYYY@{\hskip 0.5cm}YYY}
\toprule
& \multicolumn{3}{c}{\textbf{With Physical Prior}} & \multicolumn{3}{c}{\textbf{No Physical Prior}} \\
\cmidrule(lr){2-4} \cmidrule(lr){5-7}
\textbf{Level (\%)} & \textbf{Noisy} & \textbf{N2V} & \textbf{UMVD} & \textbf{Noisy} & \textbf{N2V} & \textbf{UMVD} \\
\midrule
10 \% & 33.685 $\pm$ 1.07 & 2.4096 $\pm$ 0.103 & 0.76427 $\pm$ 0.115 & 38.316 $\pm$ 14.9 & 10.016 $\pm$ 24.1 & 0.76791 $\pm$ 0.115 \\
20 \% & 130.77 $\pm$ 3.75 & 6.438 $\pm$ 0.226 & 1.6716 $\pm$ 0.191 & 130.78 $\pm$ 3.75 & 6.4515 $\pm$ 0.226 & 12.565 $\pm$ 34.4 \\
30 \% & 307.36 $\pm$ 59.5 & 12.044 $\pm$ 0.377 & 2.7006 $\pm$ 0.296 & 280.29 $\pm$ 6.27 & 12.075 $\pm$ 0.381 & 2.7239 $\pm$ 0.292 \\
\bottomrule
\end{tabularx}
\end{table}

\begin{table}[h!]
\centering
\caption{\textbf{Results of optimization with Poisson noise.} Average final MSE ($\pm$ 1 standard deviation) across ten synthetic dataset realizations for noisy, N2V, and UMVD data, comparing runs with and without physical priors.
The ground truth dataset is corrupted and denoised under decreasing number of events, $\lambda$.}
\tabtoc{Results of optimization with Poisson noise}
\label{tab:poisson_mse_split}
\small
\begin{tabularx}{\textwidth}{lYYY@{\hskip 0.5cm}YYY}
\toprule
& \multicolumn{3}{c}{\textbf{With Physical Prior}} & \multicolumn{3}{c}{\textbf{No Physical Prior}} \\
\cmidrule(lr){2-4} \cmidrule(lr){5-7}
\textbf{$\lambda$} & \textbf{Noisy} & \textbf{N2V} & \textbf{UMVD} & \textbf{Noisy} & \textbf{N2V} & \textbf{UMVD} \\
\midrule
$10^4$ & 2.6955 $\pm$ 0.0333 & 0.75475 $\pm$ 0.0401 & 0.77617 $\pm$ 0.167 & 2.6944 $\pm$ 0.0334 & 0.75418 $\pm$ 0.0408 & 0.77517 $\pm$ 0.167 \\
$10^3$ & 26.769 $\pm$ 0.375 & 4.0876 $\pm$ 0.369 & 1.3881 $\pm$ 0.163 & 26.771 $\pm$ 0.374 & 4.0872 $\pm$ 0.37 & 4.9755 $\pm$ 11.3 \\
$10^2$ & 245.73 $\pm$ 4.01 & 25.29 $\pm$ 2.59 & 5.1173 $\pm$ 0.259 & 245.77 $\pm$ 4.01 & 25.343 $\pm$ 2.59 & 5.133 $\pm$ 0.257 \\
\bottomrule
\end{tabularx}
\end{table}

\begin{table}[h!]
\centering
\caption{\textbf{Results of optimization with Impulse noise.} Average final MSE ($\pm$ 1 standard deviation) across ten synthetic dataset realizations for noisy, N2V, and UMVD data, comparing runs with and without physical priors.
The ground truth dataset is corrupted and denoised under increasing pixel ratios, $\alpha$.}
\tabtoc{Results of optimization with Impulse noise}
\label{tab:impulse_ds_mse_split}
\small
\begin{tabularx}{\textwidth}{lYYY@{\hskip 0.5cm}YYY}
\toprule
& \multicolumn{3}{c}{\textbf{With Physical Prior}} & \multicolumn{3}{c}{\textbf{No Physical Prior}} \\
\cmidrule(lr){2-4} \cmidrule(lr){5-7}
\textbf{$\alpha$} & \textbf{Noisy} & \textbf{N2V} & \textbf{UMVD} & \textbf{Noisy} & \textbf{N2V} & \textbf{UMVD} \\
\midrule
0.2 & 59.582 $\pm$ 5.45 & 20.953 $\pm$ 1.87 & 1.8039 $\pm$ 0.164 & 67.541 $\pm$ 23.5 & 27.281 $\pm$ 19.8 & 1.8028 $\pm$ 0.163 \\
0.3 & 248.6 $\pm$ 12.3 & 95.128 $\pm$ 5.64 & 3.3002 $\pm$ 0.358 & 248.63 $\pm$ 12.3 & 95.153 $\pm$ 5.63 & 3.3013 $\pm$ 0.355 \\
0.4 & 750.28 $\pm$ 16.3 & 471.45 $\pm$ 131 & 6.9205 $\pm$ 0.777 & 750.46 $\pm$ 16.2 & 325.46 $\pm$ 16.3 & 6.9256 $\pm$ 0.774 \\
\bottomrule
\end{tabularx}
\end{table}

\begin{table}[h!]
\centering
\caption{\textbf{Results of optimization with Impulse noise.} Average final mean squared error (MSE) ($\pm$ 1 standard deviation) across ten synthetic dataset realizations for noisy, N2V, and UMVD data, comparing runs with and without physical priors.
The ground truth dataset is corrupted and denoised as described in Section \ref{subsec:SI_CHR}.}
\tabtoc{Results of optimization with composite noise}
\label{tab:combined_mse_split}
\small
\begin{tabularx}{\textwidth}{YYY@{\hskip 0.5cm}YYY}
\toprule
\multicolumn{3}{c}{\textbf{With Physical Prior}} 
& \multicolumn{3}{c}{\textbf{No Physical Prior}} \\
\cmidrule(lr){1-3} \cmidrule(lr){4-6}
\textbf{Noisy} & \textbf{N2V} & \textbf{UMVD} 
& \textbf{Noisy} & \textbf{N2V} & \textbf{UMVD} \\
\midrule
106.66 $\pm$ 66.3 & 34.858 $\pm$ 2.73 & 3.2621 $\pm$ 0.221 
& 76.884 $\pm$ 4.82 & 34.867 $\pm$ 2.73 & 3.2614 $\pm$ 0.223 \\
\bottomrule
\end{tabularx}
\end{table}
 
\clearpage

\begin{figure}[H]
    \centering
    \includegraphics[width=\linewidth]{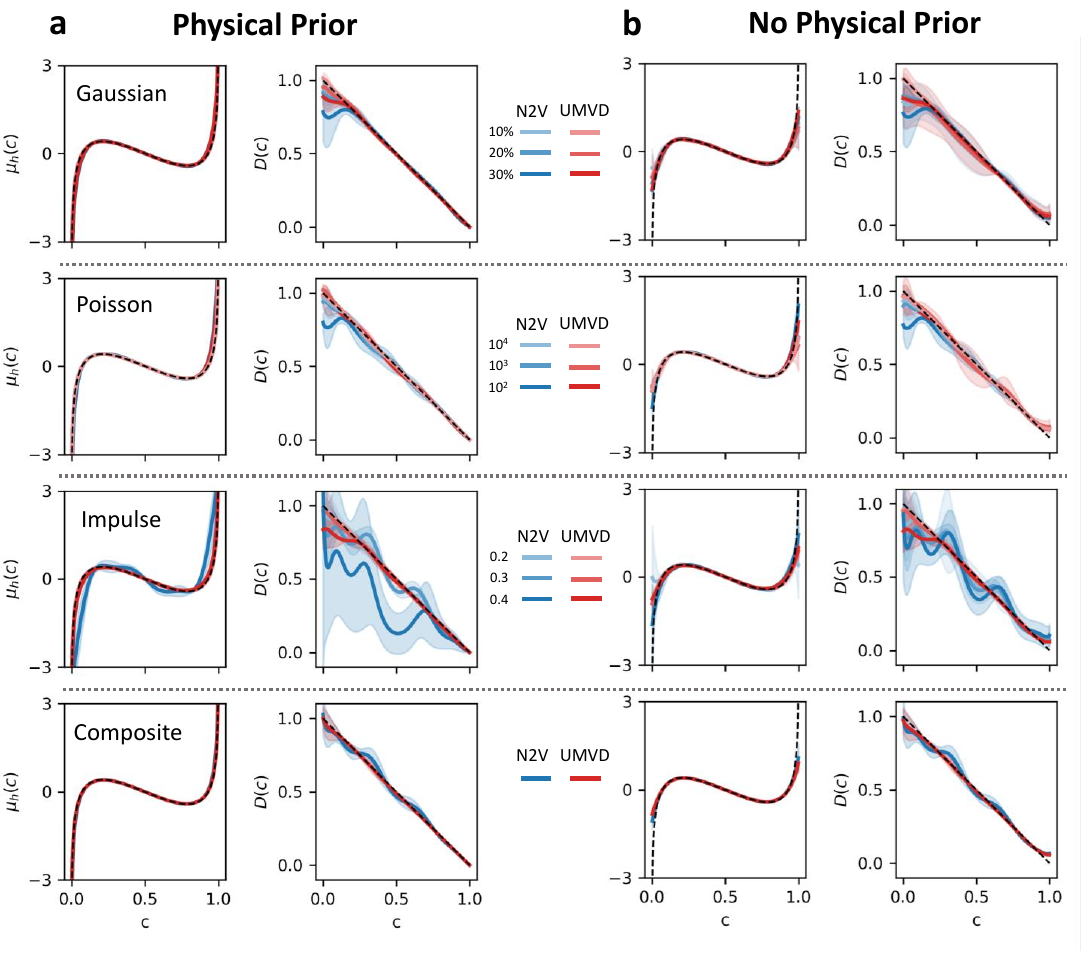}
    \caption{\textbf{Results of optimization under increasing noise intensity levels with and without physical prior on learned functions.}
    Generally, we see across all noise distributions increasing noise intensity corresponds to increasing bias and uncertainty in the learned results with similar improved performance to denoising as discussed in Figure 2 and the main text.
    The physical prior decreases the bias within the chemical potential term, especially near the end points, and it decreases the uncertainty in the learned diffusivity under higher noise levels.
    }
    \figtoc{Results of optimization under additional noise levels and priors}
    \label{fig:chr_addoptres}
\end{figure}


\subsection{STXM}\label{subsec:SI_figtab_STXM}
\begin{figure}[H]
\centering
\includegraphics[width=\textwidth]{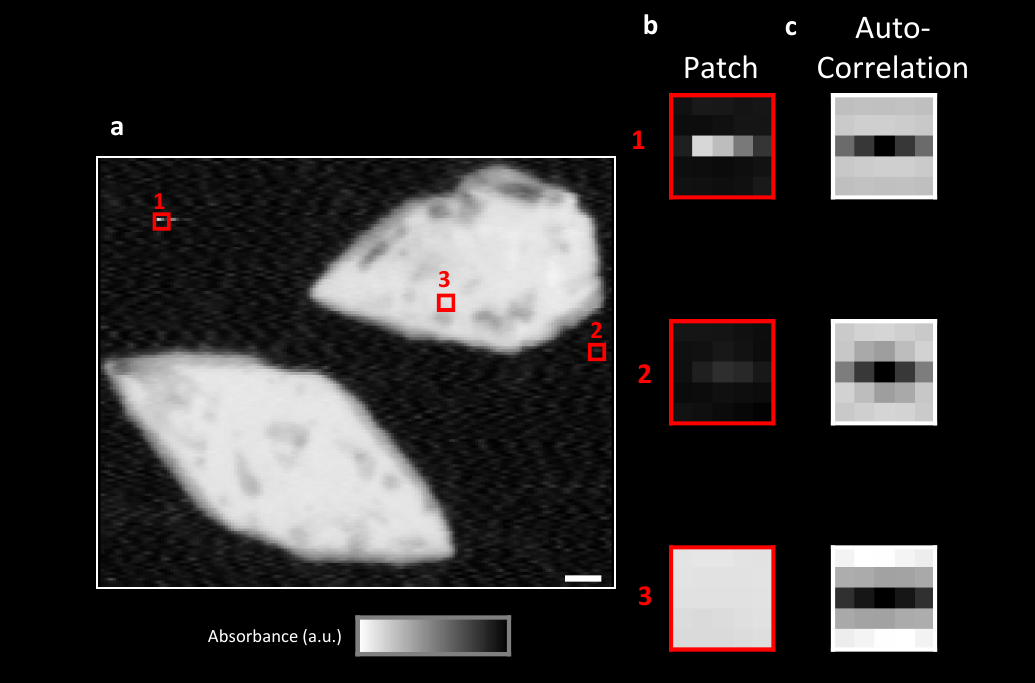}
\caption{\textbf{Spatially correlated noise in scanning transmission X-ray microscopy.}
\textbf{a}, Raw X-ray absorbance at 706 eV for the particle shown in Fig.~\ref{fig:lfp_stxm}, displaying horizontal streak artifacts across both particles and background.  
\textbf{b}, Enlarged patches highlighting (1) a beam artifact, (2) the image background, and (3) a particle region. All patches reveal noise or artifacts, most prominently in the background.  
\textbf{c}, Spatial auto-correlation computed within each patch, showing strong horizontal correlation in the measurement noise. Scale bar 500nm}
\label{fig:stxm_autocorr}
\figtoc{Spatially correlated noise in STXM}
\end{figure}

\begin{figure}[H]
\centering
\includegraphics[width=\textwidth]{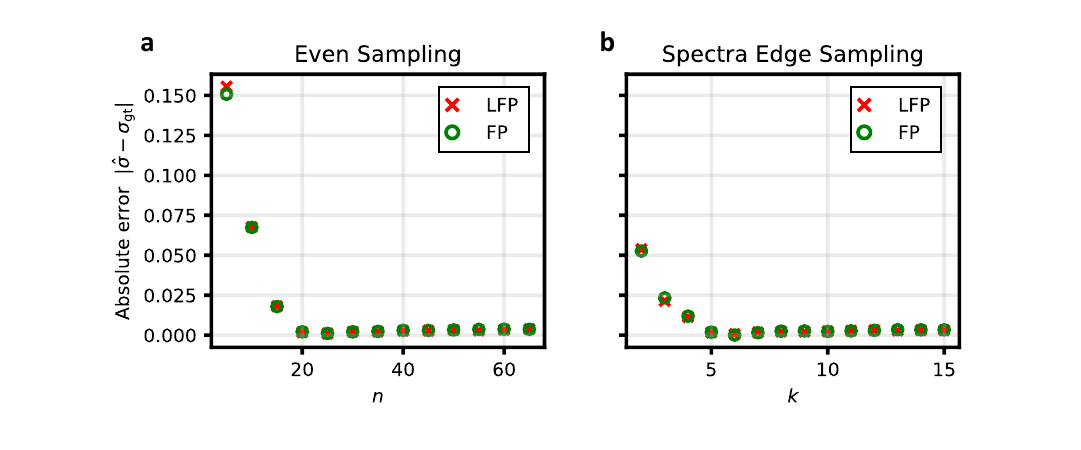}
\caption{\textbf{Convergence of lithium composition uncertainty estimates under Poisson noise using bootstrap resampling on a synthetic X-ray absorbance dataset.}  
\textbf{a}, Convergence of an even sampling strategy, where $n$ bootstrap samples are taken at linearly spaced energies between 695–715 eV.  
The estimated uncertainty in lithium composition converges at $n=20$ samples, well below the 65 samples used to estimate the uncertainty in the experimental data in this study.  
\textbf{b}, Convergence of an alternative edge-focused strategy, with one pre-edge sample and $k$ samples placed on the rising edge (705–707 eV) and falling edge (711–714 eV), giving $n=2k+1$ total samples.  
Here, the composition uncertainty converges at $k=5$ ($n=11$ samples), demonstrating more efficient use of spectra-specific information compared to uniform sampling.}
\label{fig:stxm_unc_conv}
\figtoc{Convergence of synthetic STXM uncertainty estimates}
\end{figure}

\begin{figure}[H]
\centering
\includegraphics[width=\textwidth]{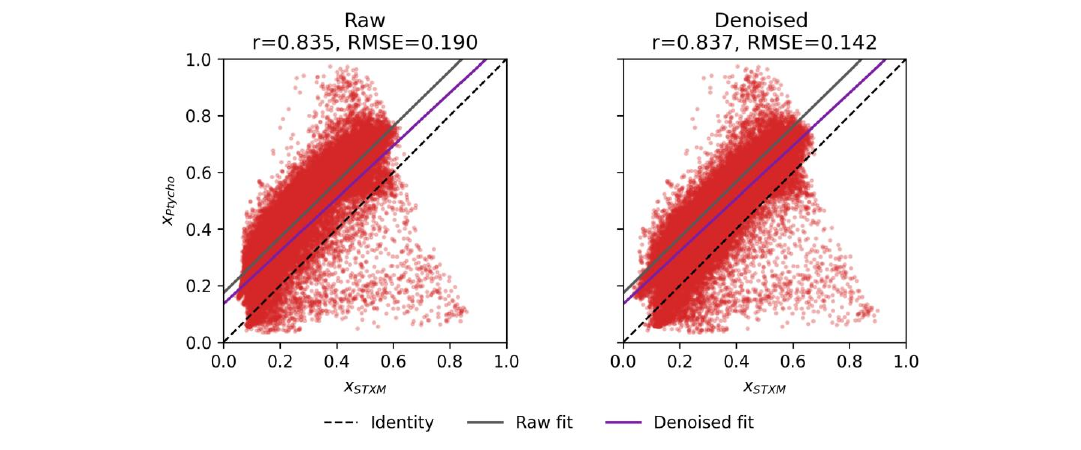}
\caption{\textbf{Correlation of X-ray absorbance measured by STXM and spectro-ptychography.}  
Pointwise correlation of X-ray absorbance ($x$) values obtained from STXM and spectro-ptychography across $n=18$ scans of 5 particles.  
Both raw and denoised STXM data show bias relative to the ptychography reference, reflected in a non-zero intercept.  
The denoised data reduces the root mean squared error (RMSE) by 25\%, indicating improved agreement with the high-resolution measurements.  
Residual bias may arise from imperfect image alignment or small differences in energy calibration between STXM and ptychography.}
\label{fig:stxm_ptychocorr}
\figtoc{Correlation of STXM and ptychography X-ray absorbance}
\end{figure}

\subsection{Optical Microscopy}\label{subsec:SI_figtab_OM}
\begin{figure}[H]
\centering
\includegraphics[width=\textwidth]{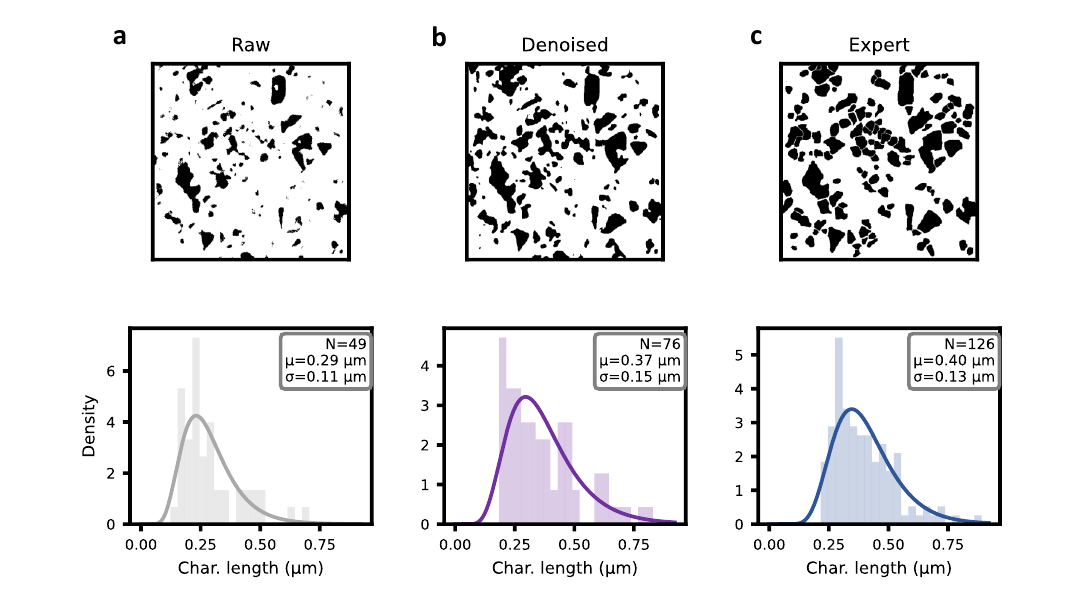}
\caption{\textbf{Particle segmentations from $k$-means clustering and corresponding log-normal size distributions.}  
\textbf{a}, Segmentation of raw images misses many small- and medium-sized particles ($N=49$ versus $N=126$ in the manual reference) and underestimates the size of detected particles.  
The resulting size distribution has a significantly smaller mean ($\mu=0.29~\mu$m compared to $\mu=0.40~\mu$m).  
\textbf{b}, Segmentation of denoised images more accurately captures the electrode morphology relative to the expert reference.  
The mean particle size is comparable, though the distribution is slightly broader because no post-processing is applied to separate primary particles within secondary agglomerates.  
\textbf{c}, Manual expert segmentation, used as ground truth for optical data analysis, yields a mean particle size of $\mu=0.40~\mu$m with standard deviation $\sigma=0.13~\mu$m.}
\label{fig:om_sizedist}
\figtoc{Optical microscopy segmentations and size distributions}
\end{figure}

\begin{figure}[H]
\centering
\includegraphics[width=.75\textwidth]{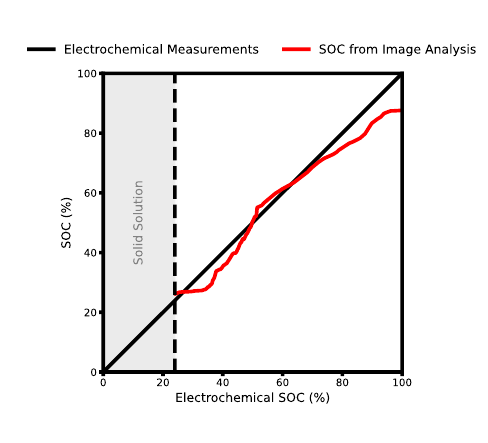}
\caption{\textbf{Comparison of state-of-charge (SOC) estimated from optical image analysis and electrochemical measurements.}  
SOC values derived from image analysis show good agreement with the electrochemical measurements.}
\label{fig:om_validation}
\figtoc{Validation of SOC from image analysis}
\end{figure}

\begin{figure}[H]
\centering
\includegraphics[width=\textwidth]{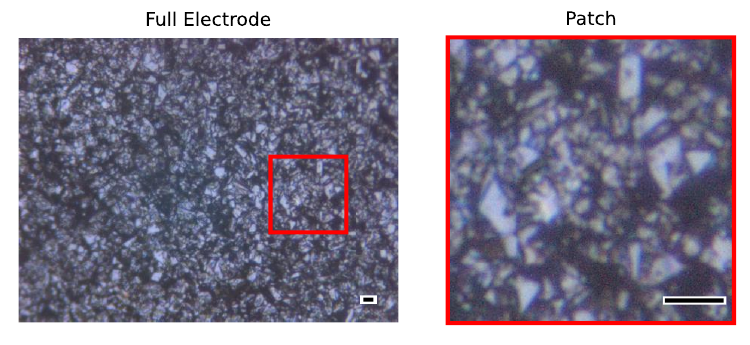}
\caption{\textbf{Optical microscopy image of the full graphite electrode.}  
The entire electrode image is used to train the denoising model, while the red inset indicates the patch used for image analysis in this study. Scale bar 10$\mu$m}
\label{fig:om_patch}
\figtoc{Selected patch for optical microscopy data anlysis}
\end{figure}

\subsection{Neutron Radiography}\label{subsec:SI_figtab_NR}

\newpage

\section{Supplementary Notes}

\subsection{Cahn-Hilliard Simulations and PDE-constrained Optimization}\label{subsec:SI_CHR}

Simulations used a concentration-dependent diffusivity $D(c) = 1 - c$ and chemical potential $\mu_h(c) = \log \frac{c}{1 - c} + \Omega (1 - 2c)$ with $\Omega = 3.0$.
A gradient penalty term \(\kappa = 2 \times 10^{-3}\) was applied, and periodic boundary conditions were used.
Initial conditions were varied to generate 10 independent realizations, each simulated on a $128 \times 128$ spatial grid with 501 time points to capture the evolution from initial spinodal decomposition to late-stage coarsening.

Gaussian noise was added to the ground-truth images with relative amplitudes of 10\%, 20\%, and 30\% (scaled to the image standard deviation).
Poisson noise was applied with $\lambda = 10^2, 10^3, 10^4$, and impulse (salt-and-pepper) noise was applied with probabilities of 0.2, 0.3, and 0.4, followed by a $3 \times 3$ median filter to suppress extreme outliers.
Composite noise combined Gaussian (10\%), Poisson ($\lambda = 10^4$), and impulse (0.2) components applied in sequence (impulse then Poisson then Gaussian), with median filtering applied last.
After corruption, pixel values were clipped to the range $[0.01, 0.99]$ to prevent numerical instabilities during optimization. The main text figure shows the worst-case noise distributions: Gaussian with 30\% relative intensity, Poisson $\lambda = 100$, and impulse with probability 0.4.

For PDE-constrained optimization, seven time points were selected during the coarsening process.
Each consecutive pair was used as the initial and final time points in a multiple-shooting procedure. 
The unknown functions were parameterized using Legendre polynomials, and positivity of the diffusivity was enforced by exponentiation.
Specifically, we learn the following functional forms:
\begin{align*}
    \text{Physical Prior: }& \mu_h(c) = \log\left(\frac{c}{1 - c}\right) + \sum^{N}_{n=1} a_nP_n(c), \ D(c) = \exp\left(\sum^{N}_{n=1} b_nP_n(c)\right) \\
    \text{No Physical Prior: }& \mu_h(c) = \sum^{N}_{n=1} a_nP_n(c), \ D(c) = \exp\left(\sum^{N}_{n=1} b_nP_n(c)\right)
\end{align*}
where $P_n(c), c\in[0,1]$ are the normalized Legendre polynomials,$a_n,b_n$ are the coefficients optimized during the learning procedure, and degree $N=11$ polynomials were used in this study.
The physical prior enforces the contribution to the free energy from configurational entropy on a lattice \cite{bazant_theory_2013}.

The forward problems were solved using \texttt{diffrax} on GPUs \cite{kidger_neural_2022}, and the parameters were optimized with the Levenberg–Marquardt algorithm including an \(L_2\) regularization term.
The full set of results and final mean squared errors (MSEs) are provided in Figure \ref{fig:chr_addoptres} and Tables \ref{tab:gaussian_mse_split},\ref{tab:poisson_mse_split},\ref{tab:impulse_ds_mse_split},\ref{tab:combined_mse_split}.
All simulations and optimizations were implemented using our JAX-based package \texttt{mosaix-pde} \cite{cohen_mosaix-pde_2025}, which provides GPU acceleration and automatic differentiation.

\subsection{Scanning Transmission X-ray Microscopy Image Processing}\label{subsec:SI_STXM}

We reprocessed raw X-ray absorbance measurements of LFP collected from previous \textit{ex situ} \cite{deng_correlative_2021} and \textit{in situ} \cite{lim_origin_2016} experiments.  
Raw images from both datasets were denoised following the procedure described in Subsection~\nameref{subsec:denoise_meth}.  
Each energy level measurement was cropped to the particle boundary to reduce background contributions to the data distribution.  
For each particle, multi-energy X-ray images were registered and aligned using phase cross-correlation \cite{guizar-sicairos_efficient_2008}.  
Absorbance measurements were converted to optical density according to  
\begin{equation*}
    OD = \log\left(\frac{I}{I_0}\right),
\end{equation*}
where $I$ is the transmitted intensity and $I_0$ is the incident intensity, estimated by Otsu thresholding of the background absorbance.  
Optical density was converted to lithium composition using the Fe-$L_3$ edge.  
For \textit{in situ} images, the optical density was normalized to the pre-edge absorption at 703 eV and used to solve the following matrix equation:  
\begin{equation*}
\begin{bmatrix}
\Delta S_{706} \\
\Delta S_{713}
\end{bmatrix}
=
\begin{bmatrix}
\Delta\mathrm{LFP}_{706} & \Delta\mathrm{FP}_{706}  \\
\Delta\mathrm{LFP}_{713} & \Delta\mathrm{FP}_{713}
\end{bmatrix}
\begin{bmatrix}
a \\
b
\end{bmatrix},
\end{equation*}
where $\Delta$ indicates normalization to the pre-edge value, $\Delta S_{706,713}$ are the normalized absorptions at 706 eV and 713 eV, $\Delta\mathrm{LFP}_{706}=0.64$ and $\Delta\mathrm{LFP}_{713}=0.11$ are the reference absorptions for lithiated $\mathrm{LiFePO}_4$, and $\Delta\mathrm{FP}_{706}=0.05$ and $\Delta\mathrm{FP}_{713}=0.60$ are the reference absorptions for delithiated $\mathrm{FePO}_4$, taken from prior work \cite{lim_origin_2016}.  
The lithium fraction was then calculated as  
\begin{equation*}
    X = \frac{a}{a + b},
\end{equation*}
where $X=1$ corresponds to fully lithiated LFP and $X=0$ to fully delithiated FP, with values clipped between 0 and 1.  
For \textit{ex situ} images, $n$ energy levels were collected, and a generalized system of equations was solved using a non-negative least-squares algorithm with reference absorption spectra for LFP and FP:  
\begin{equation*}
\begin{bmatrix}
\vdots \\
\Delta S_{n} \\
\vdots
\end{bmatrix}
=
\begin{bmatrix}
\vdots & \vdots \\ 
\Delta\mathrm{LFP}_{n} & \Delta\mathrm{FP}_{n} \\
\vdots & \vdots
\end{bmatrix}
\begin{bmatrix}
a \\
b
\end{bmatrix}.
\end{equation*}

Uncertainty in the lithium fraction was estimated using bootstrapping across energy levels.  
Previous work \cite{zhao_learning_2023} reported $\sigma_X=0.072$ using bootstrapping with $n=5$ energy levels, which we find insufficient to reproduce the true uncertainty in an extensive investigation investigation of a synthetic case study (Fig. \ref{fig:stxm_unc_conv}).
Other estimates placed the standard deviation at $\sigma_X=0.06$ by calculating the variance of lithium fraction in (de)lithiated reference particles, though this assumes perfectly uniform particles and neglects internal heterogeneity.  
To improve uncertainty quantification, we exploited reference spectra with $n=65$ energy levels, sufficient to recover the true variability in synthetic data (Fig. \ref{fig:stxm_unc_conv}), and applied bootstrapping across all available energies.  
This approach yielded $\sigma_X=0.012$ for raw images and $\sigma_X=0.014$ for denoised images, representing an 83\% reduction relative to previous estimates \cite{zhao_learning_2023}.  
Additional commentary and comparisons of raw and denoised uncertainties are provided in the Supplementary Information.  

Equivalent ptychography absorption data was obtained directly from prior work \cite{deng_correlative_2021} and converted to lithium fraction using the same procedure.  
For comparison with ptychography, the higher-resolution absorption maps were cropped to particle boundaries, resized to match the cropped STXM images, padded with the mean background absorbance, and re-aligned.  
Correlations between lithium fraction determined from STXM and ptychography are reported in Supplementary Fig. \ref{fig:stxm_ptychocorr}.
For dynamic analysis, the same registration procedure was applied to track particle composition during electrochemical cycling.  

\subsection{Optical Microscopy Image Processing}\label{subsec:SI_OM}

The video dataset of a C/60 lithiation–delithiation cycle was denoised using UMVD \cite{aiyetigbo_unsupervised_2024}, following the procedure described in Subsection~\nameref{subsec:denoise_meth}.  
Image stacks were aligned using drift correction in Nanopyx \cite{saraiva_nanopyx_2023}.  
For analysis, a representative $512 \times 512$ patch from the lithiation half-cycle was selected (Fig. \ref{fig:om_patch}).  

To segment particles algorithmically, all RGB pixels were first extracted from the spatiotemporal patch and transformed into LAB color space.  
K-means clustering ($n=4$) was applied to identify background, blue, red, and gold clusters, and each pixel was assigned to the nearest cluster.  
Non-background pixels were designated as particles.  
Expert segmentations were generated in Napari \cite{sofroniew_napari_2025}, starting from the clustering mask and refined by exploiting the fact that phase transitions proceed mosaically across individual particles \cite{zhao_population_2019}, allowing grain boundaries to be delineated by color differences.  

For phase classification, raw and denoised LAB datasets were masked using the expert segmentations, and k-means clustering ($n=4$) was applied again.  
To enforce physically consistent phase transitions, we implemented heuristic corrections.  
Specifically, all particles were set to blue in the solid-solution regime, pixels remaining blue in the final frame were reassigned to background in all frames, and after the stage II transition no pixels were allowed to remain blue.  
Transitions from gold to red were enforced as irreversible, as were transitions from blue to red, and once pixels became background they remained so for all subsequent frames.  
These corrections reflect the physical expectation that graphite progresses sequentially from stage III (blue) to stage II (red) to stage I (gold) on experimentally reasonable timescales.  

To validate the phase assignments, we estimated the state-of-charge (SOC) from image data by assuming $c_{\text{blue}}=0.26$, $c_{\text{red}}=0.55$, and $c_{\text{gold}}=1.0$, following \cite{agrawal_dynamic_2022}.  
The image-based SOC was calculated as $SOC_{\text{image}} = \sum_i c_i a_i$, where $a_i$ is the area fraction of phase $i$.  
Comparison with electrochemical SOC showed close agreement (Fig. \ref{fig:om_validation}).  

Particle size distributions were obtained by computing the characteristic length $V/A$, defined as the ratio of particle area to perimeter, under the assumption of uniform particle thickness consistent with the platelet morphology.  
Small objects were removed, individual particles were labeled, and the resulting distributions were fitted with log-normal functions (Fig. \ref{fig:om_sizedist}).  

Finally, population densities were quantified by calculating the average lithium concentration in each particle using the same weighted sum approach.  
Two-dimensional kernel density estimation was then applied to map the particle population density as a joint function of average concentration and characteristic size, as shown in Fig.~\ref{fig:graphite_optical}. 

\subsection{Neutron Radiography Data Processing}\label{subsec:NR_meth}

Raw \textit{operando} neutron radiographs were reprocessed from previously published experiments \cite{gober_high_2025}.
A $3 \times 3$ median filter was applied to the intensity images as well as the corresponding open-beam and dark-current references.
Then, denoising was performed on the three data sets following the procedure described in Subsection~\nameref{subsec:denoise_meth}.  

Normalized transmission was calculated as  
\begin{equation*}
    T(x,y,t) = \frac{I(x,y,t) - DC(x,y)}{OB(x,y) - DC(x,y)},
\end{equation*}
where $I(x,y,t)$ is the raw intensity, $DC(x,y)$ is the dark current averaged over five frames, and $OB(x,y)$ is the open beam averaged over five frames.
The field of view was cropped to the region containing the solvent-free cell, and normalized transmission was converted to the attenuation coefficient using the Beer–Lambert relation:  
\begin{equation*}
    T(x,y,t) = \exp\left(-\Sigma(x,y,t) \cdot \delta(x)\right),
\end{equation*}
where $\Sigma$ is the attenuation coefficient and $\delta(x)$ is the electrode thickness in the image plane.
The latter was determined by assuming a cylindrical geometry and computing the chord length:  
\begin{equation}
    \delta(x) = 2 \sqrt{r^2 - d^2(x)},
\end{equation}
with $r$ the cell radius and $d(x)$ the distance from the cell center to pixel $(x,y)$.
Pixel-to-distance conversion was based on a calibration of 1289.8 pixels per cm. 

Dynamic analysis was performed by dividing the attenuation data into eight half-cycles (four charge, four discharge).
Each subset was normalized to the first frame of the corresponding half-cycle using Equation~\ref{eq:att_norm}.
At the final frame of each half-cycle, Otsu thresholding \cite{otsu_threshold_1979} was applied to determine cutoffs for the anode and cathode.
Pixels above the threshold were classified as \emph{active} and used to calculate the fraction of active electrode area shown in Fig.~\ref{fig:neutron}.
The active window boundaries (green vertical lines in Fig.~\ref{fig:neutron}) were defined by the $y$-positions where the fraction of active pixels in a cross-section fell below 20\%.



\renewcommand{\bibsection}{%
    \section*{Supplementary References}%
    \phantomsection%
    \addcontentsline{toc}{section}{Supplementary References}%
  }
\putbib                         
\end{bibunit}

\end{document}